%% file: main.tex
\documentclass{article}
\usepackage[numbers, compress]{natbib} 
\usepackage[final]{neurips_2022}
\usepackage[dvipsnames]{xcolor} 
\input{macros}  

\makeatletter
\let\@fnsymbol\@arabic
\makeatother

\title{Influencing Long-Term Behavior in\\Multiagent Reinforcement Learning}

\author{
  Dong-Ki Kim\thanks{MIT-LIDS\;\;${}^{2}$IBM-Research\;\;${}^{3}$MIT-IBM Watson AI Lab\;\;${}^{4}$Mila\;\;${}^{5}$University of Oxford}
  \textsuperscript{\:,}\footnotemark[3]\\
  \texttt{dkkim93@mit.edu}
  \And
  Matthew Riemer\footnotemark[2]\textsuperscript{\:\:\:,}\footnotemark[3]\textsuperscript{\:\:\:,}\footnotemark[4]\\
  \texttt{mdriemer@us.ibm.com}
  \And
  Miao Liu\footnotemark[2]\textsuperscript{\:\:\:,}\footnotemark[3]\\
  \texttt{miao.liu1@us.ibm.com}
  \And
  Jakob N. Foerster\footnotemark[5]\\
  \texttt{jakob.foerster@eng.ox.ac.uk}
  \And
  Michael Everett\footnotemark[1]\textsuperscript{\:\:\:,}\footnotemark[3]\\
  \texttt{mfe@mit.edu}
  \And
  Chuangchuang Sun\footnotemark[1]\textsuperscript{\:\:\:,}\footnotemark[3]\\
  \texttt{ccsun1@mit.edu}
  \And
  Gerald Tesauro\footnotemark[2]\textsuperscript{\:\:\:,}\footnotemark[3]\\
  \texttt{gtesauro@us.ibm.com}
  \And
  Jonathan P. How\footnotemark[1]\textsuperscript{\:\:\:,}\footnotemark[3]\\
  \texttt{jhow@mit.edu}
}

\begin{document}
\maketitle
\vspace{-0.5cm}
\begin{abstract}
    The main challenge of multiagent reinforcement learning is the difficulty of learning useful policies in the presence of other simultaneously learning agents whose changing behaviors jointly affect the environment's transition and reward dynamics. An effective approach that has recently emerged for addressing this non-stationarity is for each agent to anticipate the learning of other agents and influence the evolution of future policies towards desirable behavior for its own benefit. Unfortunately, previous approaches for achieving this suffer from myopic evaluation, considering only a finite number of policy updates. As such, these methods can only influence transient future policies rather than achieving the promise of scalable equilibrium selection approaches that influence the behavior at convergence. In this paper, we propose a principled framework for considering the limiting policies of other agents as time approaches infinity. Specifically, we develop a new optimization objective that maximizes each agent's average reward by directly accounting for the impact of its behavior on the limiting set of policies that other agents will converge to. Our paper characterizes desirable solution concepts within this problem setting and provides practical approaches for optimizing over possible outcomes. As a result of our farsighted objective, we demonstrate better long-term performance than state-of-the-art baselines across a suite of diverse multiagent benchmark domains.
\end{abstract}
\input{introduction.tex}
\input{problem_statement.tex}
\input{method.tex}
\input{experiment.tex}
\input{related_work.tex}
\input{conclusion.tex}

\subsubsection*{Acknowledgments}
Research funded by IBM (as part of the MIT-IBM Watson AI Lab initiative).

\bibliographystyle{unsrt}
\bibliography{main}

\newpage
\input{checklist}

\newpage
\input{appendix}

\end{document}

%% file: macros.tex
\usepackage[utf8]{inputenc} 
\usepackage[T1]{fontenc}    
\usepackage{microtype}
\usepackage{graphicx}  
\usepackage{booktabs}
\usepackage{hyperref}
\usepackage{algorithm}
\usepackage{algcompatible}
\usepackage{amsmath}
\usepackage{soul}
\usepackage[authormarkuptext=name,addedmarkup=bf,authormarkupposition=left]{changes}
\usepackage{xspace}
\usepackage{dsfont}
\usepackage{bm}
\usepackage{bbm}
\usepackage{nicefrac}       
\usepackage[scaled=.8]{beramono} 
\usepackage{amsthm}
\usepackage{amsfonts}
\usepackage{amssymb}
\usepackage{array}
\usepackage{mathtools}
\usepackage{caption}
\usepackage{subcaption}
\usepackage{fontawesome}
\usepackage{thmtools}
\usepackage{thm-restate}
\usepackage{lipsum}
\usepackage{pifont}
\usepackage{makecell}
\usepackage{tablefootnote}
\usepackage[tight-spacing=true]{siunitx}
\usepackage{multirow}
\usepackage{afterpage}
\usepackage{tabularx} 
\usepackage{url}
\usepackage{tikz}
\usepackage{indentfirst}
\usepackage{cleveref}
\usepackage{enumitem}
\usepackage{epsdice}
\usepackage{wrapfig}
\usepackage[export]{adjustbox}
\usepackage{xparse}

\makeatletter
\newcommand{\subalign}[1]{%
  \vcenter{%
    \Let@ \restore@math@cr \default@tag
    \baselineskip\fontdimen10 \scriptfont\tw@
    \advance\baselineskip\fontdimen12 \scriptfont\tw@
    \lineskip\thr@@\fontdimen8 \scriptfont\thr@@
    \lineskiplimit\lineskip
    \ialign{\hfil$\m@th\scriptstyle##$&$\m@th\scriptstyle{}##$\hfil\crcr
      #1\crcr
    }%
  }%
}
\makeatother

\Crefname{figure}{Figure}{Figures}
\crefname{figure}{Figure}{Figures}
\crefname{table}{Table}{Tables}
\crefformat{table}{Table~#1}
\crefformat{equation}{equation (#1)}
\Crefformat{Equation}{Equation (#1)}
\crefformat{algorithm}{Algorithm~#1}

\hypersetup{
  colorlinks=true,
  citecolor=Blue,
  linkcolor=Blue,
  urlcolor=Blue
}

\newcommand\sdots{\hbox to 1em{.\hss.\hss.}} 
\DeclareMathAlphabet\mathbfcal{OMS}{cmsy}{b}{n} 
\newcommand{\smallsum}{\textstyle\sum\limits}
\newcommand*{\QEDB}{\null\nobreak\hfill\ensuremath{\square}}%
\DeclareMathSymbol{\shortminus}{\mathbin}{AMSa}{"39}

\setitemize[0]{leftmargin=*}

\newif\ifcomments
\commentsfalse 

\ifcomments
	\newcommand{\mr}[1]{\color{orange}MR: (#1)\color{black}\xspace}  
	\newcommand{\ml}[1]{\color{red}ML: (#1)\color{black}\xspace}  
	\newcommand{\me}[1]{\color{red}ME: (#1)\color{black}\xspace}  
	\newcommand{\dk}[1]{\color{red}DK: (#1)\color{black}\xspace}  
	\newcommand{\jf}[1]{\color{red}JF: (#1)\color{black}\xspace}  
	\newcommand{\jh}[1]{\color{red}JH: (#1)\color{black}\xspace}  
	\newcommand{\gt}[1]{\color{red}GT: (#1)\color{black}\xspace}  
\else
    \newcommand{\mr}[1]{}  
	\newcommand{\ml}[1]{}  
	\newcommand{\me}[1]{}  
	\newcommand{\dk}[1]{}  
	\newcommand{\jf}[1]{}  
	\newcommand{\jh}[1]{}  
	\newcommand{\gt}[1]{}  
	\newcommand{\cc}[1]{}  
\fi

%% file: introduction.tex
\section{Introduction}\label{sec:introduction}
Learning in multiagent reinforcement learning (MARL) is fundamentally difficult because an agent interacts with other simultaneously learning agents in a shared environment~\citep{Busoniu2010}. 
The joint learning of agents induces non-stationary environment dynamics from the perspective of each agent, requiring an agent to adapt its behavior with respect to potentially unknown changes in the policies of other agents~\citep{papoudakis19nonstationarity}.
Notably, non-stationary policies will converge to a recurrent set of joint policies by the end of learning. 
In practice, this converged joint policy can correspond to a game-theoretic solution concept, such as a Nash equilibrium~\citep{Nash48} or more generally a cyclic correlated equilibrium~\citep{zinkevich06cyclic}, but multiple equilibria can exist for a single game with some of these Pareto dominating others~\citep{Nowe2012}. 
Hence, a critical question in addressing this non-stationarity is how individual agents should behave to influence convergence of the recurrent set of policies towards more desirable limiting behaviors. 

Our key idea in this work is to consider the limiting policies of other agents as time approaches infinity.
Specifically, the converged behavior of this dynamic multiagent system is not due to some arbitrary stochastic processes, but rather each agent's underlying learning process that also depends on the behaviors of the other interacting agents.
As such, effective agents should model how their actions can influence the limiting behavior of other agents and leverage those dependencies to shape the convergence process.
This farsighted perspective contrasts with recent work that also considers influencing the learning of other agents~\citep{foerster17lola,letcher2018stable,xie20lili,kim21metamapg,wang2021influencing,lu2022modelfree}. 
While these approaches show improved performance over methods that neglect the learning of other agents entirely~\citep{lowe17maddpg,forester17coma,iqbal19masac}, they suffer from myopic evaluation: only considering a few updates to the policies of other agents or optimizing for the discounted return, which only considers a finite horizon time of $1/(1\!-\!\gamma)$ for discount factor $\gamma$~\citep{kearns2002near}.

\paragraph{Our contribution.} With this insight, we make the following primary contributions in this paper:
\begin{itemize}[leftmargin=*, wide, labelindent=0pt, topsep=0pt]
    \itemsep 0in 
    \item \textbf{Formalization of multiagent non-stationarity (\Cref{sec:active-markov-game}).} We introduce an active Markov game setting that formalizes MARL with simultaneously learning agents as a directed graphical model and captures the underlying non-stationarity over time. We detail how such a system eventually converges to a stationary periodic distribution. As such, the objective is to maximize its long-term rewards over this distribution and, if each agent maximizes this objective, the resulting multiagent system settles into a new and general equilibrium concept that we call an active equilibrium.
    \item \textbf{Practical framework for optimizing an active Markov game (\Cref{sec:method}).} 
    We outline a practical approach for optimization in this setting, called FUlly Reinforcing acTive influence witH averagE Reward (FURTHER). Our approach is based on a policy gradient and Bellman update rule tailored to active Markov games. Moreover, we show how variational inference can be used to approximate the update function of other agents and support decentralized execution and training. 
    \item \textbf{Comprehensive evaluation of our approach (\Cref{sec:evaluation}).} We demonstrate that our method consistently converges to a more desirable limiting distribution than baseline methods that either neglect the learning of others~\citep{iqbal19masac} or consider their learning with a myopic perspective~\citep{xie20lili} in various multiagent benchmark domains. 
    We also demonstrate that FURTHER provides a flexible framework such that it can incorporate recent advances in multiagent learning and improve performance in large-scale settings by leveraging the mean-field method~\citep{yang18mean-field-marl}.
\end{itemize}

%% file: problem_statement.tex
\section{Problem Statement: Active Markov Game}\label{sec:active-markov-game}
This work studies a general multiagent learning setting, where each agent interacts with other independently learning agents in a shared environment. 
Agents in this setting update their policies based on recent experiences which are affected by the joint actions of all agents.
As such, while an agent cannot directly modify the future policies of other interacting agents, the agent can actively influence them by changing its own actions.
In this section, we first formalize the presence of this causal influence in multiagent interactions by introducing the new framework of an active Markov game. 
We then formalize solution concepts and objectives for learning within this framework.
Finally, we discuss dependence on initial states and policies, detailing choices that we can make to minimize the impact of these initial conditions on behavior after convergence.

\begin{figure}[h]
    \vskip-0.07in
    \centering
    \includegraphics[height=3.3cm]{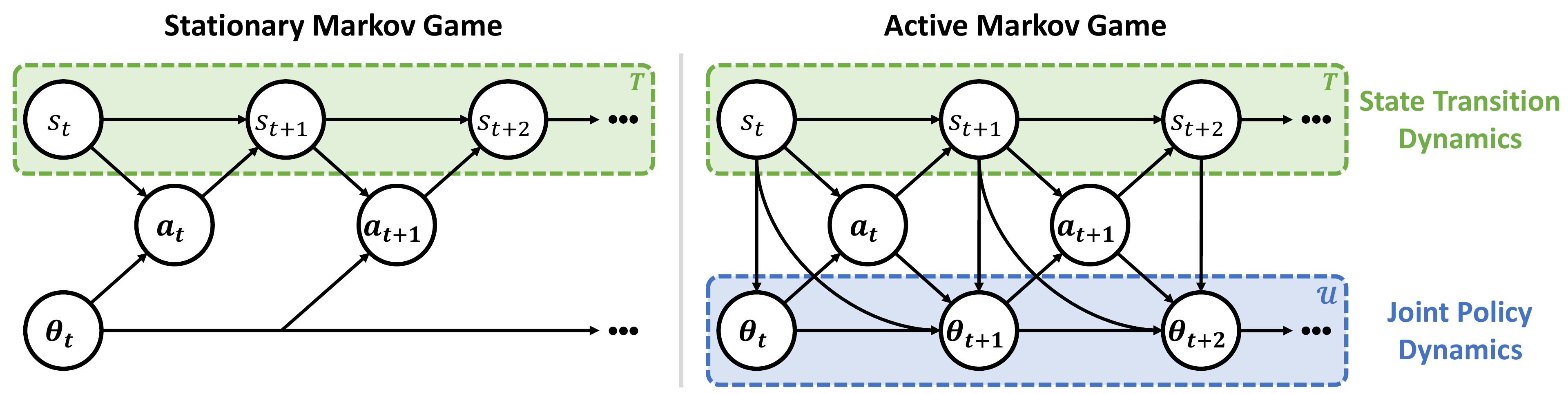}
    \caption{Within the stationary Markov game setting, agents wrongly assume that other agents will have stationary policies into the future. In contrast, agents in an active Markov game recognize that other agents have non-stationary policies based on the Markovian update functions.}
    \label{fig:active-markov-game}
    \vskip-0.15in
\end{figure}

\subsection{Directed Graphical Model of Active Markov Game}
We define an active Markov game as a tuple $\mathcal{M}_n\!=\!\langle\mathbfcal{I}, \mathcal{S},\mathbfcal{A},\mathcal{T},\mathbfcal{R},$ $\bm{\Theta},\mathbfcal{U}\rangle$;
$\mathbfcal{I}\!=\!\{1,\sdots,n\}$ is the set of $n$ agents;
$\mathcal{S}$ is the state space;
$\mathbfcal{A}\!=\!\times_{i \in \mathcal{I}} \mathcal{A}^{i}$ is the joint action space;
$\mathcal{T}\!:\!\mathcal{S}\!\times\!\mathbfcal{A}\!\mapsto\!\mathcal{S}$ is the state transition function;
$\mathbfcal{R}\!=\!\times_{i \in \mathcal{I}} \mathcal{R}^{i}$ is the joint reward function; 
$\bm{\Theta}\!=\!\times_{i \in \mathcal{I}} \Theta^{i}$ is the joint policy parameter space; and
$\bm{\mathcal{U}}\!=\!\times_{i \in \mathcal{I}} \mathcal{U}^{i}$ is the joint Markovian policy update function.
We typeset sets in bold for clarity.
Compared to the stationary Markov game that effectively represents MARL with wrongly assumed stationary policies in the future, the active Markov game considers how policies change over time (see~\Cref{fig:active-markov-game}).
Specifically, at each timestep $t$, each agent $i$ executes an action at a current state $s_{t}\!\in\!\mathcal{S}$ according to its stochastic policy $a^i_{t}\!\sim\!\pi^{i}(\cdot|s_{t};\theta^i_{t})$ parameterized by $\theta^{i}_{t}\!\in\!\Theta^{i}$. 
A joint action $\bm{a}_{\bm{t}}\!=\!\{a^{i}_t,\bm{a}^{\bm{\shortminus i}}_{\bm{t}}\}$ yields a transition from $s_{t}$ to $s_{t+1}$ with probability $\mathcal{T}(s_{t+1}|s_{t},\bm{a}_{\bm{t}})$, where the notation $\bm{\shortminus i}$ indicates all other agents except agent $i$.
Each agent $i$ then obtains a reward according to its reward function $r^{i}_t\!=\!\mathcal{R}^i(s_t,\bm{a}_{\bm{t}})$ and updates its policy parameters according to $\mathcal{U}^{i}(\theta^{i}_{t+1}|\theta^{i}_{t},\tau^{i}_{t})$, where $\tau^{i}_{t}\!=\!\{s_{t},\bm{a}_{\bm{t}},r^{i}_{t},s_{t+1}\}$ denotes agent $i$'s transition. 
This process continues until the convergence of non-stationary policies.
Notably, the joint policy update function $\bm{\mathcal{U}}$ is a function of $a^{i}_{t}$, which affects the state transitions and rewards, so agent $i$ can actively influence future joint policies by changing its own behavior. 
Modeling this influence rather than ignoring it is the main advantage of using active Markov games rather than the stationary Markov game formalism.

\subsection{Solution Concepts in Active Markov Games}
The formalism of active Markov games provides a principled framework for each agent to model the impact of its behavior on joint future policies.
In this section, we study the theoretical convergence properties of an active Markov game and develop relevant terminology that will help us characterize this convergence. 
We begin by formalizing the limiting behavior as a stationary periodic distribution.

\textbf{Definition 1.} (Stationary $k$-Periodic Distribution). \textit{The limiting behavior of an active Markov game can be represented by a stationary periodic probability distribution over the joint space of states and policies, defined as a stationary conditional distribution with respect to a period of order $k$}:
\begin{align}\label{eqn:stationary-periodic-distribution}
\begin{split}
\mu_{k}(s,\bm{\theta}|s_0,\bm{\theta_0},\ell)=p(s_t\!=\!s,\bm{\theta_t}\!=\!\bm{\theta}|s_0,\bm{\theta_{0}},\ell) \quad\forall t\!\geq\!0,s,s_0\!\in\!\mathcal{S},\bm{\theta},\bm{\theta_0}\!\in\!\bm{\Theta},
\end{split}
\end{align}
\textit{where $\ell\!=\!t\%k$ with \% denoting the modulo operation. The stationary $k$-periodic distribution satisfies the following property as its time averaged expectation stays stationary in the limit}:
\begin{align}\label{eqn:stationary-periodic-distribution-property}
\begin{split}
\frac{1}{k}\smallsum_{\ell=1}^{k}&\mu_k(s_{\ell+1},\bm{\theta}_{\bm{\ell+1}}|s_0,\bm{\theta_0},\ell\!+\!1)\!=\!\frac{1}{k}\smallsum_{\ell=1}^{k}\smallsum_{s_{\ell},\bm{\theta_{\ell}}}\mu_k(s_{\ell},\bm{\theta_{\ell}}|s_0,\bm{\theta_0},\ell)\smallsum_{\bm{a_\ell}}\bm{\pi}(\bm{a_\ell}|s_{\ell};\bm{\theta_{\ell}})\\
&\quad\mathcal{T}(s_{\ell+1}|s_{\ell},\bm{a_{\ell}})\;\bm{\mathcal{U}}(\bm{\theta_{\ell+1}}|\bm{\theta_{\ell}},\bm{\tau_{\ell}})\quad\forall s_{\ell+1} \!\in\!\mathcal{S},\bm{\theta_{\ell+1}}\!\in\!\bm{\Theta}.
\end{split}
\end{align}
Our notion of a stationary $k$-periodic distribution provides a flexible representation for characterizing the limiting distribution, generalizing from fully stationary fixed-point convergence (when $k\!=\!1$) to the extreme case of totally non-stationary convergence (when $k\!\rightarrow\!\infty$).

Having defined the joint convergence behavior of an active Markov game, we can now develop an objective that each agent can optimize to maximize its long-term rewards.
Our key finding is that the average reward formulation, developed for single-agent learning~\citep{puterman94mdp,sutton98rlbook}, is well suited for studying the limiting behavior of other interacting agents in multiagent learning. 
In particular, the average reward formulation maximizes the agent's average reward per step with equal weight given to immediate and delayed rewards, unlike the discounted return objective.
Once the joint policy converges to the stationary periodic distribution, rewards collected by this recurrent set of policies govern each agent's average reward as $t\!\rightarrow\!\infty$. 
Thus, optimizing for the average reward in an active Markov game encourages agents to consider how to influence the limiting set of policies after convergence rather than transient policies that are only experienced momentarily.

\textbf{Definition 2.} (Active Average Reward Objective). \textit{
Each agent $i$ in an active Markov game aims to find policy parameters $\theta^{i}$ and update function $\mathcal{U}^{i}$ that maximize its expected average reward $\rho^{i}\!\in\!\mathbb{R}$}:
\begin{align}\label{eqn:average-reward-objective}
\begin{split}
\max_{\theta^{i},\,\mathcal{U}^{i}}\rho^{i}(s,\bm{\theta},\bm{\mathcal{U}})\!:&=\!\max_{\theta^{i},\,\mathcal{U}^{i}}\lim_{T\rightarrow\infty}\!\mathbb{E}\Big[\frac{1}{T}\!\smallsum_{t=0}^{T}\!\mathcal{R}^{i}(s_t,\bm{a}_{\bm{t}})\Big|\substack{s_{0}=s,\;\bm{\theta}_{\bm{0}}=\bm{\theta},\\\bm{a_{0:T}}\sim\bm{\pi}(\bm{\cdot}|s_{0:T};\bm{\theta}_{\bm{0:T}}),\\s_{t+1}\sim\mathcal{T}(\cdot|s_{t},\bm{a_t}),\bm{\theta_{t+1}}\sim\bm{\mathcal{U}}(\cdot|\bm{\theta_{t}},\bm{\tau_{t}})}\Big]\\
&=\!\max_{\theta^{i},\,\mathcal{U}^{i}}\frac{1}{k}\smallsum_{\ell=1}^{k}\smallsum_{s_{\ell},\bm{\theta_{\ell}}}\!\!\mu_k(s_{\ell},\bm{\theta_{\ell}}|s,\bm{\theta},\ell)\smallsum_{\bm{a_\ell}}\bm{\pi}(\bm{a_\ell}|s_{\ell};\bm{\theta_{\ell}})\mathcal{R}^{i}(s_{\ell},\bm{a_{\ell}}),
\end{split}
\end{align}
where $T$ denotes the time horizon. It is important to note that \Cref{eqn:average-reward-objective} has no preference over the large equivalence class of update functions that eventually converge to an optimal limiting behavior, and we only require finding an update function in this class even if the convergence rate is slow. This is advantageous for our discussion to come about solution concepts in active Markov games. However, in our practical approach to optimization, we also optimize over the transient distribution, pushing towards solutions with lower regret by modeling our value function based on the differential returns as in Proposition 2.
We also note that even if a single agent maximizes this objective, agents will not necessarily arrive at any kind of equilibrium. 
This is because other agents may have sub-optimal or biased update functions beyond the agent's control, and a rational agent can potentially seek to converge to an average reward that is better for it than that of any equilibrium as a result.
Additionally, whether an agent just seeks to optimize its policy or maximize its update function as well depends on the kind of solution concept that is desired. 
For example, finding a fixed stationary policy equates to using an update function that arrives at a fixed point, whereas we can also optimize over the update function if we seek to find an optimal non-stationary policy as in the meta-learning literature~\citep{kim21metamapg,alshedivat2018continuous}.

If all agents maximize the active average reward objective, we arrive at a new and general equilibrium concept that we call an active equilibrium, where no agents can further optimize its average reward:

\textbf{Definition 3.} (Active Equilibrium). \textit{In an active Markov game, an active equilibrium is joint policy parameters $\bm{\theta^{*}}\!=\!\{\theta^{i*},\bm{\theta^{\shortminus i*}}\}$ with associated joint update function $\bm{\mathcal{U}^{*}}\!=\!\{\mathcal{U}^{i*},\bm{\mathcal{U}^{\shortminus i*}}\}$ such that}:
\begin{align}
\begin{split}
\rho^{i}(s,\theta^{i*},\bm{\theta}^{\bm{\shortminus i*}},\mathcal{U}^{i*},\bm{\mathcal{U}^{\shortminus i*}})\!\geq\!\rho^{i}(s,\theta^{i},\bm{\theta}^{\bm{\shortminus i*}},\mathcal{U}^{i},\bm{\mathcal{U}^{\shortminus i*}}) \quad\forall i\!\in\!\mathcal{I},s\!\in\!\mathcal{S},\theta^{i}\!\in\!\Theta^{i},\mathcal{U}^{i}\!\in\!\mathbb{U}^{i}.
\end{split}
\end{align}
where $\mathbb{U}^{i}$ denotes the space of agent $i$'s update functions. 
Our active equilibrium definition is related to non-stationary solution concepts in game theory, such as the non-stationary Nash equilibrium~\citep{daskalakis22gametheory}, that search for a sequence of best-response joint policies. However, these non-stationary solutions are generally intractable to compute due to the unconstrained sequence over the infinite horizon and the resulting large policy search space size. By contrast, the active equilibrium provides a more refined and practical notion than these solution concepts by having a constraint on the sequence based on the update functions. We also note the generality of active equilibrium that it can correspond to other standard solution concepts as we impose restrictions on relevant variables:

\textbf{Remark 1.} (Connection to Existing Solution Concepts). \textit{Stationary Nash~\cite{Nash48} and correlated equilibria \citep{correlated87} are special kinds of active equilibria when $k\!=\!1$ and joint action distributions are independent and correlated, respectively. Cyclic Nash and cyclic correlated equilibria \cite{zinkevich06cyclic} are also special cases of an active equilibrium if $k\!>\!1$, the joint update function is deterministic,} and joint action distributions are independent and correlated, respectively.

\subsection{Addressing Sensitivity to Initial Conditions}\label{sec:stochastically-stable-distribution}
The recurrent set of converged joint policies is generally dependent on initial states and policies, as specified by the conditioned initial variables in~\Cref{eqn:stationary-periodic-distribution,eqn:stationary-periodic-distribution-property,eqn:average-reward-objective}.
This initial condition dependence implies that there can be instances of poor convergence performance simply due to undesirable initial states and policies (see~\Cref{sec:multichain-example} for an example).
In this paper, we address this sensitivity to initial conditions by considering the stochastically stable periodic distribution, which is a special case of the stationary periodic distribution. 
The stochastic distribution describes the limiting joint behavior when each agent has communicating strategies (i.e., for every pair of policy parameters $\theta^{i},\theta^{i\prime}\!\in\!\Theta^{i}$, $\theta^{i}$ transitions to $\theta^{i\prime}$ in a finite number of steps with non-zero probability and vice versa) by adding noise $\epsilon$ to its update function $\mathcal{U}^{i}_{\epsilon}$, and noise $\epsilon\!\rightarrow\!0$ over time (i.e., $\lim_{t\rightarrow\infty}\mathcal{U}^{i}_{\epsilon}\!=\!\mathcal{U}^{i}$).
Importantly, the stochastic distribution provides an important analytical benefit of independent convergence with respect to the initial conditions. 
Specifically, assuming communicating state transitions $\mathcal{T}$, if only agent $i$'s update function is perturbed, then we arrive at the notion of self-stable periodic distribution:

\textbf{Definition 4.} (Self-Stable Periodic Distribution). \textit{Given communicating state transition $\mathcal{T}$, if noise $\epsilon$ is added only to the agent $i$'s update function $\mathcal{U}^{i}_{\epsilon}$, we achieve the stationary $k$-periodic distribution independent of the initial state and the agent $i$'s initial policy as $\epsilon\rightarrow 0$ over time}:
\begin{align}
\begin{split}
\vspace{-0.2cm}
\frac{1}{k}\smallsum_{\ell=1}^{k}&\mu_k(s_{\ell+1},\bm{\theta}_{\bm{\ell+1}}|\bm{\theta^{\shortminus i}_0},\ell\!+\!1)\!=\!\frac{1}{k}\smallsum_{\ell=1}^{k}\smallsum_{s_{\ell},\bm{\theta_{\ell}}}\mu_k(s_{\ell},\bm{\theta_{\ell}}|\bm{\theta^{\shortminus i}_0},\ell)\smallsum_{\bm{a_\ell}}\bm{\pi}(\bm{a_\ell}|s_{\ell};\bm{\theta_{\ell}})\\
&\;\mathcal{T}(s_{\ell+1}|s_{\ell},\bm{a_{\ell}})\;\bm{\mathcal{U}}(\bm{\theta_{\ell+1}}|\bm{\theta_\ell},\bm{\tau_{\ell}})\quad\forall s_{\ell+1}\!\in\!\mathcal{S},\bm{\theta_{\ell+1}}\!\in\!\bm{\Theta}.
\end{split}
\end{align}
Similarly, if the full joint update function is perturbed with noise $\bm{\mathcal{U}_\epsilon}$, this induces a unique stationary periodic distribution independent of the initial state and initial joint policy:

\textbf{Definition 5.} (Jointly-Stable Periodic Distribution). \textit{Given communicating state transition $\mathcal{T}$, if noise $\epsilon$ is added to the joint update function $\bm{\mathcal{U}_{\epsilon}}$, we achieve the same stationary $k$-periodic distribution independent of the initial state and the initial policies as $\epsilon\!\rightarrow\!0$ over time}:
\begin{align}
\begin{split}
\frac{1}{k}\smallsum_{\ell=1}^{k}&\mu_k(s_{\ell+1},\bm{\theta}_{\bm{\ell+1}}|\ell\!+\!1)\!=\!\frac{1}{k}\smallsum_{\ell=1}^{k}\smallsum_{s_{\ell},\bm{\theta_{\ell}}}\mu_k(s_{\ell},\bm{\theta_{\ell}}|\ell)\smallsum_{\bm{a}_{\bm{\ell}}}\bm{\pi}(\bm{a_\ell}|s_{\ell};\bm{\theta_{\ell}})\\
&\;\mathcal{T}(s_{\ell+1}|s_{\ell},\bm{a_{\ell}})\;\bm{\mathcal{U}}(\bm{\theta_{\ell+1}}|\bm{\theta_{\ell}},\bm{\tau_{\ell}})\quad\forall s_{\ell+1}\!\in\!\mathcal{S},\bm{\theta_{\ell+1}}\!\in\!\bm{\Theta}.
\end{split}
\end{align}

\textbf{Proposition 1.} (Uniqueness of Jointly-Stable Periodic Distribution). \textit{
Given communicating state transition $\mathcal{T}$ and perturbed joint update function with noise $\bm{\mathcal{U}_{\epsilon}}$, the jointly-stable periodic distribution is unique as $\epsilon\!\rightarrow\!0$ over time.}

\textit{Proof.} \hspace{0.2em}See~\Cref{sec:proof-stochastically-stable-equilibrium} for details.\QEDB

The jointly-stable periodic distribution is induced in many cases of interest to multiagent learning, including when all policies employ update functions leveraging the Greedy in the Limit with Infinite Exploration (GLIE) property~\citep{sutton98rlbook}: 1) all state-action pairs are visited infinitely often and 2) as $t\!\rightarrow\!\infty$, the behavior policy converges to the greedy policy. 
\begin{wrapfigure}[12]{r}{0.4\linewidth}
\vskip-0.15in
\centering
\includegraphics[height=3.3cm]{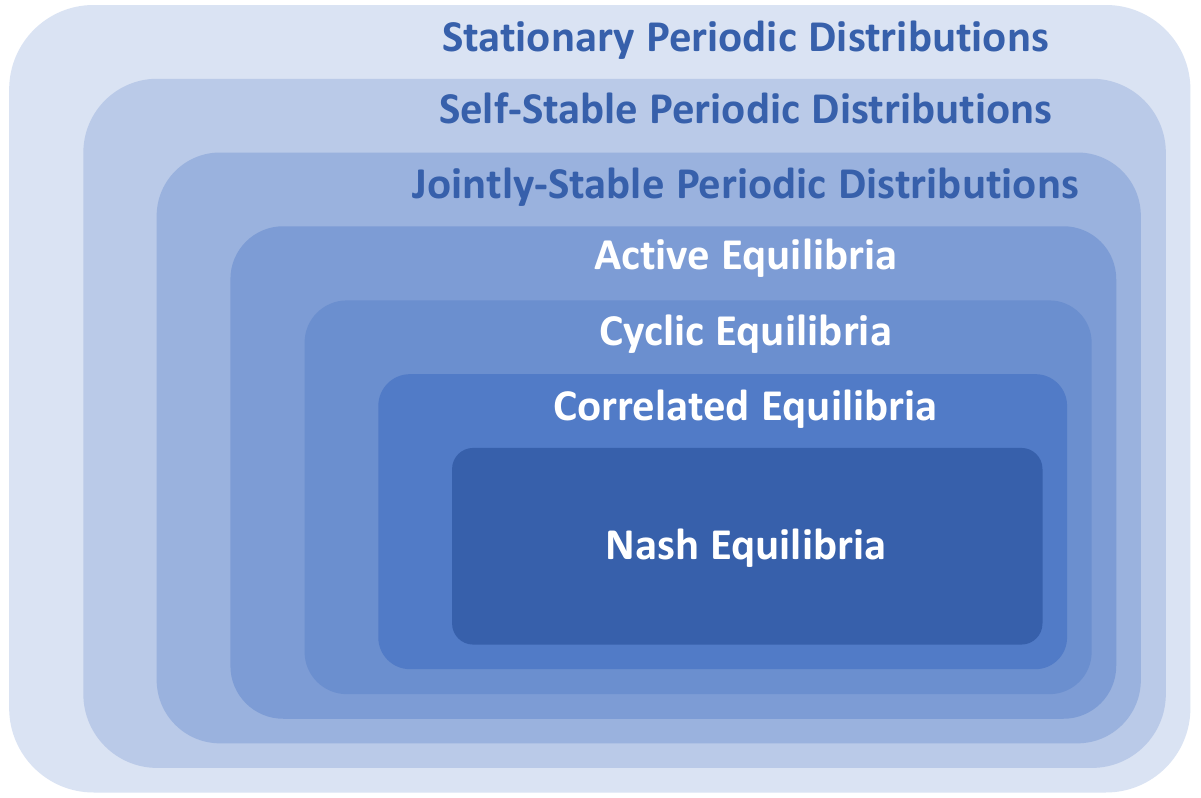}
\vskip-0.065in
\caption{Venn diagram describing relationships between the proposed distributions and equilibrium concepts.}
\label{fig:venn-diagram}
\end{wrapfigure}
In particular, a broad class of action exploration or noisy policy update functions lead to this kind of distribution~\citep{foster90stochastic,freidlin2012random,chasparis12stochastic,chasparis19perturb,wicks05ssd}.
Indeed, MARL algorithms generally rely on persistent exploration and thus satisfy GLIE.
Lastly, as demonstrated in~\Cref{fig:venn-diagram}, although maximizing over the space of jointly-stable periodic distributions confines the search space of stationary periodic distributions, the best possible active equilibria still lie within this smaller space while also allowing for optimization robust to initial conditions.
We focus on designing a learning algorithm that can find an equilibrium in the practical and confined search space of the jointly-stable distributions in the following section.

%% file: method.tex
\section{FURTHER: Practical Method for Solving Active Markov Game}\label{sec:method}
In this section, we develop a practical method, called FURTHER, for learning beneficial policies in the space of the jointly-stable periodic distributions. 
We first outline a practical version of the average reward objective and derive its policy gradient.
We then detail our model-free implementation that builds on top of soft actor-critic~\citep{haarnoja18sac} and variational inference~\citep{blei17variational} to learn policies that efficiently optimize for the average reward objective in a decentralized manner. 

\subsection{Formulation and Policy Gradient Theorem of FURTHER}\label{sec:average-reward-formulation}
While the objective in~\Cref{eqn:average-reward-objective} ideally maximizes over the space of update functions and learns a non-stationary policy, addressing the computational difficulty of long horizon meta-learning still remains an active area of research~\citep{kim21metamapg,imaml,deleu2022continuous}. 
As such, in FURTHER, we take a practical step forward and learn the optimal fixed point policy that influences joint policy behavior to maximize its long-term average reward $\rho^{i}_{\theta^i}\!\in\!\mathbb{R}$ at a state $s\!\in\!\mathcal{S}$ and policy parameters of other agents $\bm{\theta}^{\bm{\shortminus i}}\!\in\!\bm{\Theta^{\shortminus i}}$:
\begin{align}\label{eqn:further-objective}
\begin{split}
\max_{\theta^{i}}\rho^{i}_{\theta^i}(s,\bm{\theta}^{\bm{\shortminus i}})\!:=\!\max_{\theta^{i}}\!\!\lim_{T\rightarrow\infty}\!\!\mathbb{E}\Big[\frac{1}{T}\!\smallsum_{t=0}^{T}\!\mathcal{R}^{i}(s_t,\bm{a}_{\bm{t}})\Big|\substack{s_{0}=s,\;\bm{\theta}^{\bm{\shortminus i}}_{\bm{0}}=\bm{\theta}^{\bm{\shortminus i}},\\a^{i}_{0:T}\sim\pi(\cdot|s_{0:T};\theta^{i}),\bm{a}^{\bm{\shortminus i}}_{\bm{0:T}}\sim\pi(\cdot|s_{0:T};\bm{\theta}^{\bm{\shortminus i}}_{\bm{0:T}}),\\s_{t+1}\sim\mathcal{T}(\cdot|s_{t},\bm{a_{t}}),\bm{\theta^{\shortminus i}_{t+1}}\sim\bm{\mathcal{U}^{\shortminus i}}(\cdot|\bm{\theta^{\shortminus i}_{t}},\bm{\tau^{\shortminus i}_{t}})}\Big],
\end{split}
\end{align}
where the subscript $\theta^{i}$ notation denotes the implicit dependence on the learning of agent $i$'s fixed stationary policy.
As discussed in \Cref{sec:stochastically-stable-distribution}, a useful result under the jointly-stable periodic distribution is that the average reward becomes independent of the initial states and policies:
\begin{align}\label{eqn:average-reward-independent}
\begin{split}
\rho^{i}_{\theta^i}(s,\bm{\theta}^{\bm{\shortminus i}})=\rho^{i}_{\theta^i}(s',\bm{\theta}^{\bm{\shortminus i\prime}})=\rho^{i}_{\theta^i} \quad \forall s\!\neq\! s', \bm{\theta}^{\bm{\shortminus i}}\!\neq\!\bm{\theta}^{\bm{\shortminus i\prime}}.
\end{split}
\end{align}
We now derive the Bellman equation in an active Markov game that defines the relationship between the value function and average reward.

\textbf{Proposition 2.} (Active Differential Bellman Equation). \textit{
The differential value function $v^{i}_{\theta^{i}}$ represents the expected total difference between the accumulated rewards from $s$ and $\bm{\theta}^{\bm{\shortminus i}}$ and the average reward $\rho^{i}_{\theta^{i}}$~\citep{sutton98rlbook}. The differential value function inherently includes the recursive relationship with respect to $v^{i}_{\theta^{i}}$ at the next state $s^{\prime}$ and the updated policies of other agents $\bm{\theta}^{\bm{\shortminus i\prime}}$}:
\begin{align}\label{eqn:learning-aware-bellman-v}
\begin{split}
v^{i}_{\theta^{i}}(s,\bm{\theta}^{\bm{\shortminus i}})&=
\lim_{T\rightarrow\infty}\mathbb{E}\Big[\smallsum_{t=0}^{T}\big(\mathcal{R}^{i}(s_t,\bm{a}_{\bm{t}})-\rho^{i}_{\theta^{i}}\big)\Big|\;\substack{s_{0}=s,\;\bm{\theta}^{\bm{\shortminus i}}_{\bm{0}}=\bm{\theta}^{\bm{\shortminus i}},\\a^{i}_{0:T}\sim\pi(\cdot|s_{0:T};\theta^{i}),\bm{a}^{\bm{\shortminus i}}_{\bm{0:T}}\sim\bm{\pi}(\bm{\cdot}|s_{0:T};\bm{\theta}^{\bm{\shortminus i}}_{\bm{0:T}}),\\s_{t+1}\sim\mathcal{T}(\cdot|s_{t},\bm{a_{t}}),\bm{\theta^{\shortminus i}_{t+1}}\sim\bm{\mathcal{U}^{\shortminus i}}(\cdot|\bm{\theta^{\shortminus i}_{t}},\bm{\tau^{\shortminus i}_{t}})}\;\Big]\\
&=\!\smallsum_{a^{i}}\pi(a^{i}|s;\theta^{i})\!\smallsum_{\bm{a}^{\bm{\shortminus i}}}\bm{\pi}(\bm{a}^{\bm{\shortminus i}}|s;\bm{\theta}^{\bm{\shortminus i}})\!\smallsum_{s'}\mathcal{T}(s'|s,\bm{a})\!\smallsum_{\bm{\theta}^{\bm{\shortminus i\prime}}}\bm{\mathcal{U}}{}^{\bm{\shortminus i}}(\bm{\theta}^{\bm{\shortminus i\prime}}|\bm{\theta}^{\bm{\shortminus i}},\bm{\tau^{\shortminus i}})\\& 
\quad\Big[
\mathcal{R}^{i}(s,\bm{a})-\rho^{i}_{\theta^{i}}+v^{i}_{\theta^{i}}(s',\bm{\theta}^{\bm{\shortminus i\prime}})\Big].
\raisetag{15pt}
\end{split}
\end{align}
\textit{Proof.} \hspace{0.2em}See~\Cref{sec:bellman-proof} for a derivation.\QEDB

Finally, we derive the policy gradient based on the active differential Bellman equation:

\textbf{Proposition 3.} (Active Average Reward Policy Gradient Theorem). \textit{
The gradient of active average reward objective in~\Cref{eqn:further-objective} with respect to agent $i$'s policy parameters $\theta^i$ is}:
\begin{gather}\label{eqn:learning-aware-pg}
\nabla_{\theta^{i}}J^{i}_{\pi}(\theta^{i})\!=\!\frac{1}{k}\!\smallsum_{\ell=1}^{k}\smallsum_{s_{\ell},\bm{\theta^{\shortminus i}_{\ell}}}\!\mu_{k,\theta^{i}}(s_{\ell},\bm{\theta_{\ell}}|\ell)\!\smallsum_{a^{i}_{\ell}}\nabla_{\theta^{i}}\pi(a^{i}_{\ell}|s_{\ell};\theta^{i})\!\smallsum_{\bm{a_{\ell}}^{\bm{\shortminus i}}}\bm{\pi}(\bm{a_{\ell}}^{\bm{\shortminus i}}|s_{\ell};\bm{\theta_{\ell}}^{\bm{\shortminus i}})q^{i}_{\theta^{i}}(s_{\ell},\bm{\theta_{\ell}}^{\bm{\shortminus i}},\bm{a_{\ell}}),\\
\text{with}\quad\!\!\!\! q^{i}_{\theta^{i}}(s_{\ell},\bm{\theta_{\ell}}^{\bm{\shortminus i}},\bm{a_{\ell}})\!=\!\!\smallsum_{s_{\ell+1}}\!\!\mathcal{T}(s_{\ell+1}|s_{\ell},\bm{a_{\ell}})\!\!\smallsum_{\bm{\theta_{\ell+1}}^{\bm{\shortminus i}}}\!\!\bm{\mathcal{U}^{\shortminus i}}(\bm{\theta_{\ell+1}}^{\bm{\shortminus i}}|\bm{\theta_{\ell}}^{\bm{\shortminus i}},\bm{\tau_{\ell}}^{\bm{\shortminus i}})\Big[\mathcal{R}^{i}(s_{\ell},\bm{a_{\ell}})\!-\!\rho^{i}_{\theta^{i}}\!+\!v^{i}_{\theta^{i}}(s_{\ell+1},\bm{\theta_{\ell+1}}^{\bm{\shortminus i}})\Big].\nonumber
\end{gather}
\textit{Proof.} \hspace{0.2em}See~\Cref{sec:policy-gradient-proof} for a derivation.\QEDB

\subsection{Soft Actor-Critic Implementation with Variational Inference}\label{sec:sac-with-vi}
\paragraph{Algorithm overview.}
FURTHER broadly consists of inference and reinforcement learning modules. 
In practice, each agent has partial observations about others and cannot directly observe their true policy parameters $\bm{\theta^{\shortminus i}}$ and policy dynamics $\bm{\mathcal{U}^{\shortminus i}}$. 
The inference learning module predicts this hidden information about other agents leveraging variational inference~\citep{blei17variational} modified for sequential prediction. 
The inferred information becomes the input to the reinforcement learning module, which extends the policy gradient theorem in \Cref{eqn:learning-aware-pg} and learns active average reward policies sample efficiently by building on the multiagent soft actor-critic (MASAC) framework~\citep{iqbal19masac,haarnoja18sac,christodoulou2019soft}. 
We note that each agent interacts and learns these modules by only observing the actions of other agents, so our implementation supports decentralized execution and training. 
We provide further details, including implementation for $k\!>\!1$ and psuedocode, in~\Cref{sec:implementation-details}.

For simplicity, we consider the period $k\!=\!1$ and develop corresponding soft reinforcement learning optimizations in~\Cref{eqn:soft-q-value-optimization,eqn:soft-v-value-optimization,eqn:soft-policy-optimization}.

\paragraph{Inference learning module.} 
This module aims to infer the current policies of other agents and their learning dynamics based on an approximate variational inference~\citep{blei17variational}.
Specifically, we optimise a tractable evidence lower bound (ELBO), defined together with an encoder $p(\bm{\hat{z}}^{\bm{\shortminus i}}_{\bm{t+1}}|\bm{\hat{z}}^{\bm{\shortminus i}}_{\bm{t}},\tau^{i}_{t};\phi^{i}_{\text{enc}})$ and a decoder $p(\bm{a}^{\bm{\shortminus i}}_{\bm{t}}|s_{t},\bm{\hat{z}}^{\bm{\shortminus i}}_{\bm{t}};\phi^{i}_{\text{dec}})$, parameterised by $\phi^{i}_{\text{enc}}$ and $\phi^{i}_{\text{dec}}$, respectively:
\begin{align}\label{eqn:elbo}
\begin{split}
\!\!\!\!\!\!\!\!\!\mathcal{J}^{i}_{\text{elbo}}\!=\!\mathbb{E}_{p(\tau^{i}_{0:t}),p(\bm{\hat{z}}^{\bm{\shortminus i}}_{\bm{1:t}}|\tau^{i}_{0:t\!-\!1};\phi^{i}_{\text{enc}})}\!\Big[\!\smallsum_{t^{\prime}=1}^{t}\underbrace{\log p(\bm{a}^{\bm{\shortminus i}}_{\bm{t^{\prime}}}|s_{t^{\prime}},\bm{\hat{z}}^{\bm{\shortminus i}}_{\bm{t^{\prime}}};\phi^{i}_{\text{dec}})}_{\text{Reconstruction loss}}\!-\!\underbrace{D_{\text{KL}}\!\big(p(\bm{\hat{z}}^{\bm{\shortminus i}}_{\bm{t^{\prime}}}|\tau^{i}_{t^{\prime}\!-\!1};\phi^{i}_{\text{enc}})||p(\bm{\hat{z}}^{\bm{\shortminus i}}_{\bm{t^{\prime}\!-\!1}})\big)}_{\text{KL divergence}}\!\Big],
\raisetag{25pt}
\end{split}
\end{align}
where latent strategies $\bm{\hat{z}^{\shortminus i}_{t}}$ represents inferred policy parameters of other agents $\bm{\theta^{\shortminus i}_{t}}$, the encoder represents the policy dynamics of other agents $\bm{\mathcal{U}^{\shortminus i}}$ with parameters $\phi^{i}_{\text{enc}}$, and $\tau^{i}_{0:t}\!=\{\tau^{i}_{0},...,\tau^{i}_{t}\}\!$ denotes $i$'s transitions up to timestep $t$. 
We refer to~\Cref{sec:elbo-derivation} for a detailed ELBO derivation.
By optimizing the reconstruction term, the encoder aims to infer accurate next latent strategies of other agents. 
Also, by imposing the prior through the KL divergence, where we set the prior to the previous posterior with initial prior $p(\bm{\hat{z}}^{\bm{\shortminus i}}_{\bm{0}})\!=\!\mathcal{N}(0, I)$, the inferred policies from the encoder are encouraged to be sequentially consistent across time (i.e., no abrupt changes in policies of others). 

\paragraph{Reinforcement learning module.}
This module aims to learn a policy that can maximize the agent's average reward based on the inferred information about other agents.
Each agent maintains its policy $\pi(\cdot|s,\bm{\hat{z}}^{\bm{\shortminus i}};\theta^{i})$ parameterized by $\theta^{i}$, two $q$-functions $q^{i}_{\theta^{i}}(s,\bm{\hat{z}}^{\bm{\shortminus i}},\bm{a};\psi^{i}_{1})$ and $q^{i}_{\theta^{i}}(s,\bm{\hat{z}}^{\bm{\shortminus i}},\bm{a};\psi^{i}_{2})$ parameterized by $\psi^{i}_{1},\psi^{i}_{2}$, and learnable average reward $\rho^{i}_{\theta^i}\!\in\!\mathbb{R}$. 
We train the q-functions and $\rho^{i}_{\theta^i}$ by minimizing the soft Bellman residual:
\begin{align}\label{eqn:soft-q-value-optimization}
\begin{split}
\!\!\!\!\!J^{i}_{q}(\psi^{i}_{\beta},\rho^{i}_{\theta^i})\!=\!\mathbb{E}_{(s,\bm{\hat{z}}^{\bm{\shortminus i}},\bm{a},r^{i},s^{\prime},\bm{\hat{z}}^{\bm{\shortminus i\prime}})\sim\mathcal{D}^{i}}\!\Big[\!\Big(y\!-\!q^{i}_{\theta^{i}}\!(s,\bm{\hat{z}}^{\bm{\shortminus i}},\bm{a};\psi^{i}_{\beta})\Big)\!^2\!\Big]\!,
\text{ } y\!=\!r^{i}\!\!-\!\rho^{i}_{\theta^i}\!+\!v^{i}_{\theta^{i}}\!(s',\bm{\hat{z}}^{\bm{\shortminus i\prime}};\bar{\psi}^{i}_{\beta}),
\raisetag{15pt}
\end{split}
\end{align}
where $\beta\!=\!1,2$, $\mathcal{D}^{i}$ denotes $i$'s replay buffer, and $\bar{\psi}^{i}_{\beta}$ denotes the target $q$-network parameters.
The soft value function $v^{i}_{\theta^{i}}$ calculates the state value with the policy entropy $\mathcal{H}$ and entropy weight $\alpha$:
\begin{align}\label{eqn:soft-v-value-optimization}
\begin{split}
\!\!\!\!v^{i}_{\theta^{i}}\!(s,\bm{\hat{z}}^{\bm{\shortminus i}};\psi^{i})\!=\!\!\smallsum_{a^{i}}\!\pi(a^{i}|s,\bm{\hat{z}}^{\bm{\shortminus i}};\theta^{i})\!\!\smallsum_{\bm{a}^{\bm{\shortminus i}}}\!\pi(\bm{a}^{\bm{\shortminus i}}|s;\bm{\hat{z}}^{\bm{\shortminus i}})\!\!\min\limits_{\beta=1,2}\!q^{i}_{\theta^{i}}\!(s,\bm{\hat{z}}^{\bm{\shortminus i}}\!,\bm{a};\psi^{i}_{\beta})\!+\!\alpha\mathcal{H}(\pi(\cdot|s,\bm{\hat{z}}^{\bm{\shortminus i}};\theta^{i})).
\raisetag{20pt}
\end{split}
\end{align}
Finally, the policy is trained to maximize:
\begin{align}\label{eqn:soft-policy-optimization}
\begin{split}
J^{i}_{\pi}(\theta^{i})\!=\!\mathbb{E}_{(s,\bm{\hat{z}}^{\bm{\shortminus i}},\bm{a}^{\bm{\shortminus i}})\sim\mathcal{D}^{i}}\!\Big[\smallsum_{a^{i}}\pi(a^{i}|s,\bm{\hat{z}}^{\bm{\shortminus i}};\theta^{i})\!\!\min\limits_{\beta=1,2}q^{i}_{\theta^{i}}(s,\bm{\hat{z}}^{\bm{\shortminus i}},\bm{a};\psi^{i}_{\beta})\!+\!\alpha\mathcal{H}(\pi(\cdot|s,\bm{\hat{z}}^{\bm{\shortminus i}};\theta^{i}))\Big].
\end{split}
\end{align}
We note that \Cref{eqn:soft-v-value-optimization,eqn:soft-policy-optimization} are for discrete action space, and we detail optimizations for continuous action space in \Cref{sec:implementation-details}.

\paragraph{Mean-Field FURTHER.} FURTHER provides a flexible framework such that it can easily integrate recent advances in multiagent learning.
For example, by reconstructing and predicting the mean action and latent strategy of neighbor agents in~\Cref{eqn:elbo}, we can incorporate the mean-field framework to improve performance in large-scale learning settings. 
\Cref{sec:implementation-details} details the mean-field version of FURTHER with pseudocode.

%% file: experiment.tex
\section{Evaluation}\label{sec:evaluation}
We demonstrate FURTHER’s efficacy on a diverse suite of multiagent benchmark domains.
We refer to~\Cref{sec:experiment-details} for experimental details and hyperparameters. 
The code is available at \url{https://bit.ly/3fXArAo}, and video highlights are available at \url{https://bit.ly/37IWeb9}.
The mean and 95\% confidence interval computed for 20 seeds are shown in each figure.

\begin{figure}[t!]
\captionsetup[subfigure]{skip=0pt, aboveskip=13pt}
    \begin{subfigure}[b]{0.36\linewidth}
        \begin{subfigure}[b]{0.49\linewidth}
            \footnotesize
            \centering
            \setlength{\tabcolsep}{2pt}
            \begin{tabular}[b]{cc|cc}
            \multicolumn{2}{c}{} & \multicolumn{2}{c}{}\\
            \parbox[t]{2mm}{\multirow{3}{*}{}} 
                &       & $B$       & $S$       \\\cline{2-4}
            \rule{0pt}{10pt}    & $B$   & $(2,1)$ & $(0,0)$  \\
                & $S$   & $(0,0)$  & $(1,2)$ 
            \end{tabular}
            \vskip-0.13in
            \caption{Bach/Stravinsky}
            \label{fig:ibs-domain}
        \end{subfigure}
        \begin{subfigure}[b]{0.49\linewidth}
            \footnotesize
            \centering
            \setlength{\tabcolsep}{2pt}
            \begin{tabular}[b]{cc|cc}
            \multicolumn{2}{c}{} & \multicolumn{2}{c}{}\\
            \parbox[t]{2mm}{\multirow{3}{*}{\rotatebox[origin=c]{90}{}}} 
                &       & $U$       & $D$       \\\cline{2-4}
            \rule{0pt}{10pt}    & $U$   & $(4,4)$ & $(0,0)$  \\
                & $D$   & $(0,0)$  & $(8,8)$ 
            \end{tabular}
            \vskip-0.13in
            \caption{Coordination}
            \label{fig:ic-domain}
        \end{subfigure}
        \begin{subfigure}[b]{\linewidth}
            \footnotesize
            \centering
            \setlength{\tabcolsep}{2pt}
            \begin{tabular}[b]{cc|cc}
            \multicolumn{2}{c}{} & \multicolumn{2}{c}{}\\
            \parbox[t]{2mm}{\multirow{3}{*}{\rotatebox[origin=c]{90}{}}} 
                &       & $H$       & $T$       \\\cline{2-4}
            \rule{0pt}{10pt}    & $H$   & $(1,\shortminus 1)$ & $(\shortminus 1,1)$  \\
                & $T$   & $(\shortminus 1,1)$  & $(1,\shortminus 1)$ 
            \end{tabular}
            \vskip-0.13in
            \caption{Matching Pennies}
            \label{fig:imp-domain}
        \end{subfigure}
    \end{subfigure}
    \begin{subfigure}[b]{0.315\linewidth}
        \centering
        \includegraphics[height=2.65cm]{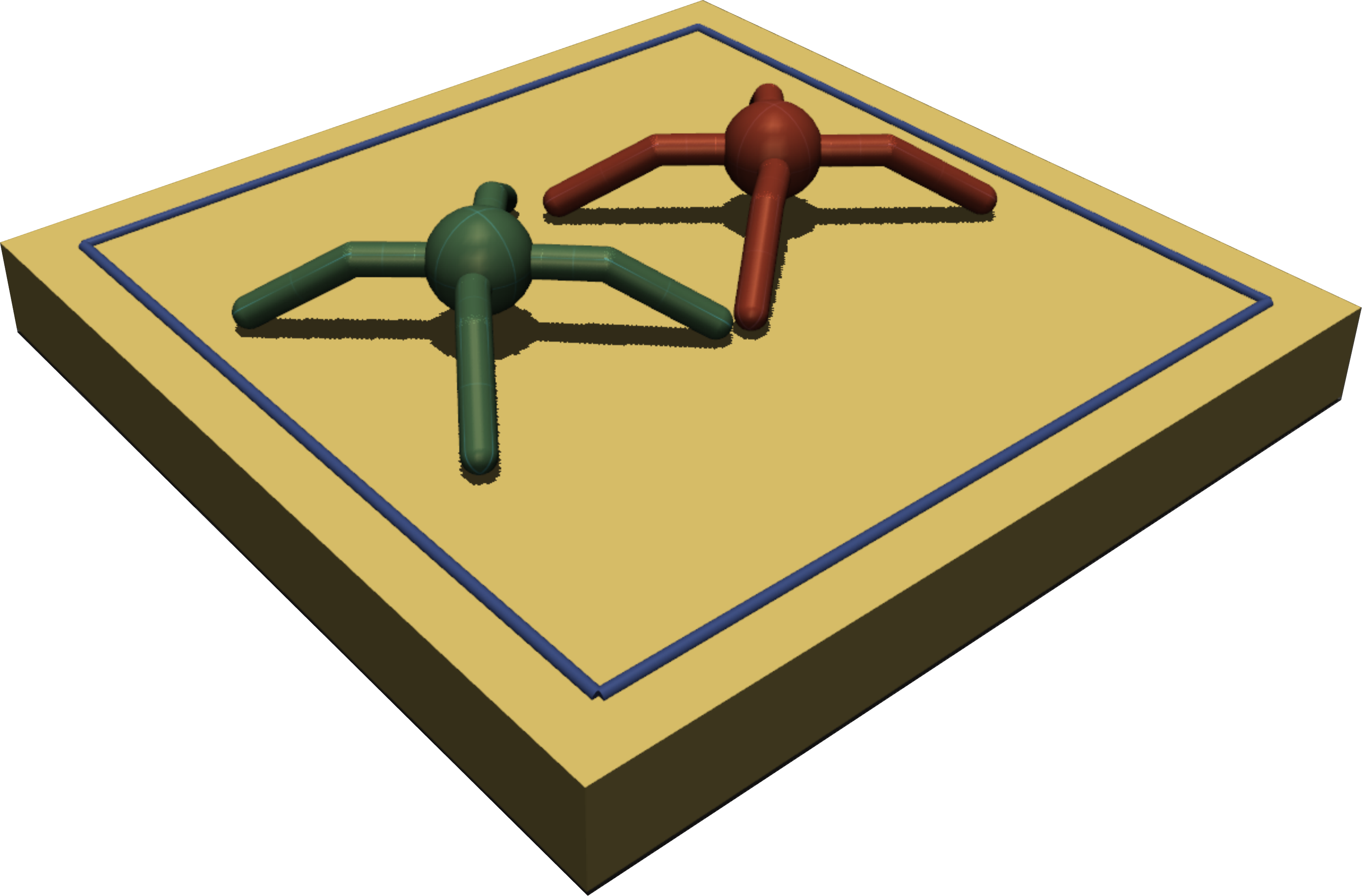}
        \vskip-0.1in
        \caption{MuJoCo RoboSumo}
        \label{fig:robosumo-domain}
    \end{subfigure}
    \begin{subfigure}[b]{0.315\linewidth}
        \centering
        \includegraphics[height=3.05cm]{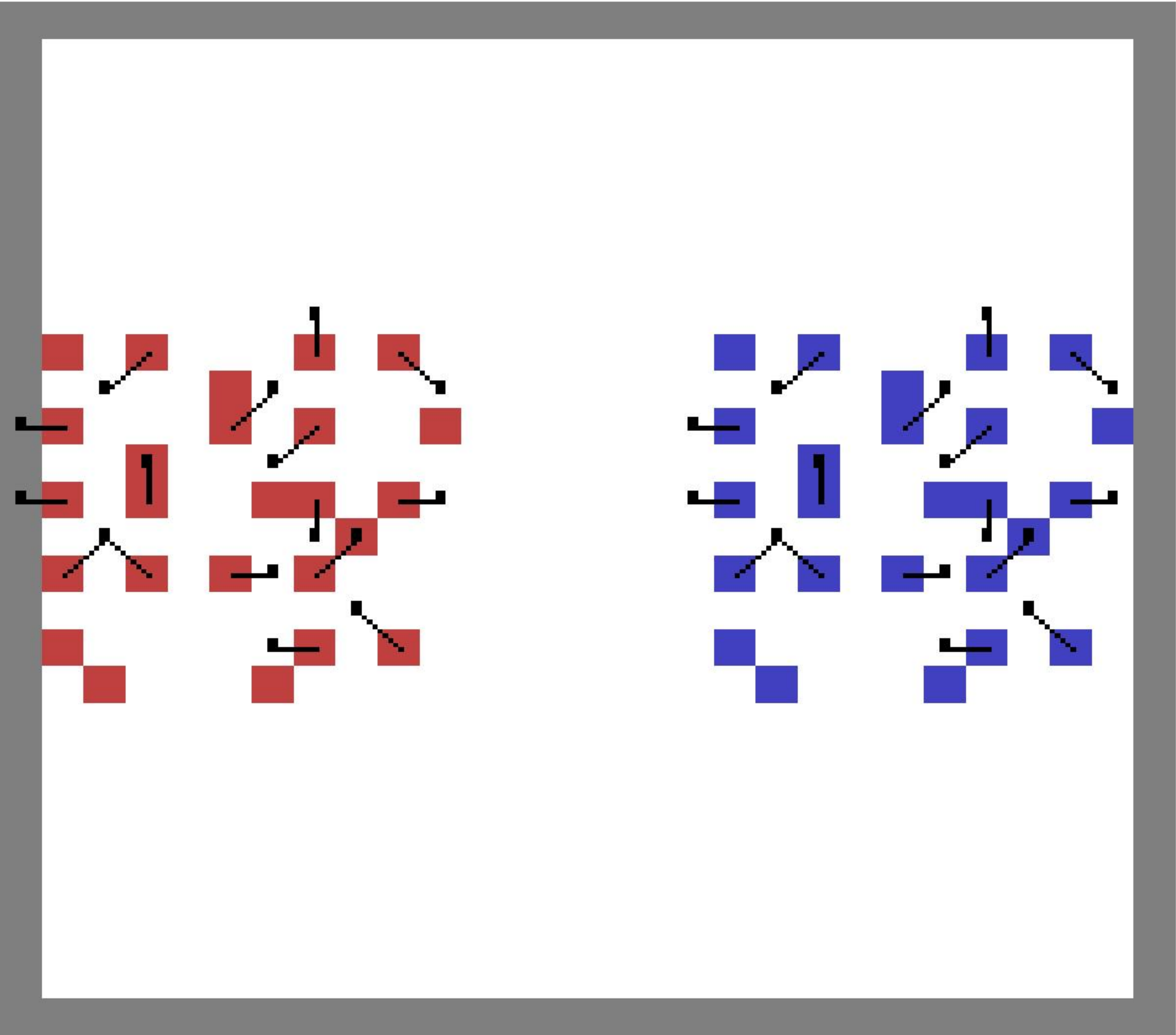}
        \vskip-0.1in
        \caption{MAgent Battle}
        \label{fig:battle-domain}
    \end{subfigure}
    \caption{\textbf{(a)-(c)} Payoff tables for Bach or Stravinsky (general-sum), coordination (cooperative), and matching pennies (competitive) games. \textbf{(d)} A competitive RoboSumo domain~\cite{alshedivat2018continuous} with two agents fighting each other. \textbf{(e)} A mixed cooperative-competitive battle domain~\cite{zheng2018magent} with 25 vs 25 agents.}
    \vskip-0.15in
\end{figure}

\paragraph{Baselines.} We compare FURTHER with the following baselines (see \Cref{appendix:baseline-details} for details):
\begin{itemize}[leftmargin=*, wide, labelindent=0pt, topsep=0pt]
    \itemsep0em 
    \item \textbf{LILI~\citep{xie20lili}:} An approach that considers the learning dynamics of other agents but suffers from myopic evaluation bias by optimizing the discounted return objective (see~\Cref{eqn:lili-objective}).
    \item \textbf{MASAC~\citep{iqbal19masac}:} An approach that extends SAC~\citep{haarnoja18sac} to a multiagent learning setting by having centralized critics~\citep{lowe17maddpg}. This baseline assumes other agents will have stationary policies in the future and thus neglects their learning (see~\Cref{eqn:masac-objective}).
\end{itemize}
We note that these selected baselines are closely related to FURTHER, optimizing different objectives with respect to $\bm{\mathcal{U}^{\shortminus i}}$. In particular, LILI and MASAC optimize the discounted return objective with and without modeling $\bm{\mathcal{U}^{\shortminus i}}$, respectively. As such, our baseline choices enable us to separately analyze the effect of FURTHER's novel average reward objective. For completeness, we also consider additional baselines of an opponent modeling method (DRON) and an incentive MARL method (MOA). These results are shown in~\Cref{appendix:additional-evaluation}.

\textbf{Question 1.} \textit{How do methods perform when playing against a $q$-learning agent?}

We consider playing the iterated Bach or Stravinsky game (IBS; see~\Cref{fig:ibs-domain}). 
This general-sum game involves conflicting elements with two pure strategy Nash equilibria, where convergence to (B,B) and (S,S) equilibrium are more preferable from agent $i$'s and $j$'s perspective, respectively. 
Suppose agent $i$ plays against a naive learner $j$, such as $q$-learner~\citep{Watkins92q-learning}, whose initial $q$-values are set to prefer action (S). 
In this experimental setting, it is ideal for agent $i$ to change $j$'s influence behavior to select (B) such that they converge to $i$'s optimal pure strategy Nash equilibrium of (B,B). 

\begin{figure}[t!]
\captionsetup[subfigure]{skip=0pt, aboveskip=3pt}
    \begin{subfigure}[b]{0.33\linewidth}
        \centering
        \includegraphics[height=3.5cm]{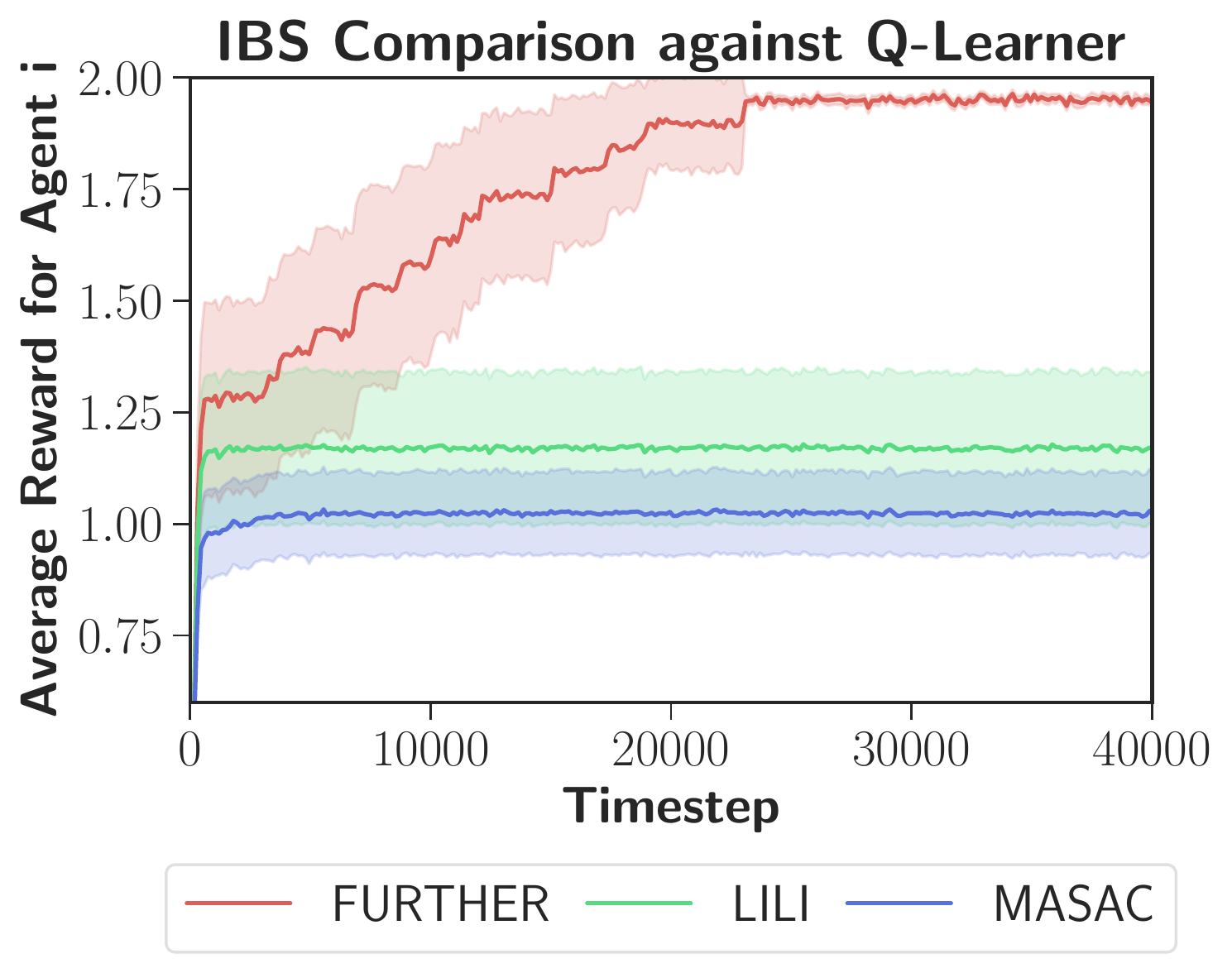}
        \caption{}
        \label{fig:ibs-result}
    \end{subfigure}
    \begin{subfigure}[b]{0.33\linewidth}
        \centering
        \includegraphics[height=3.5cm]{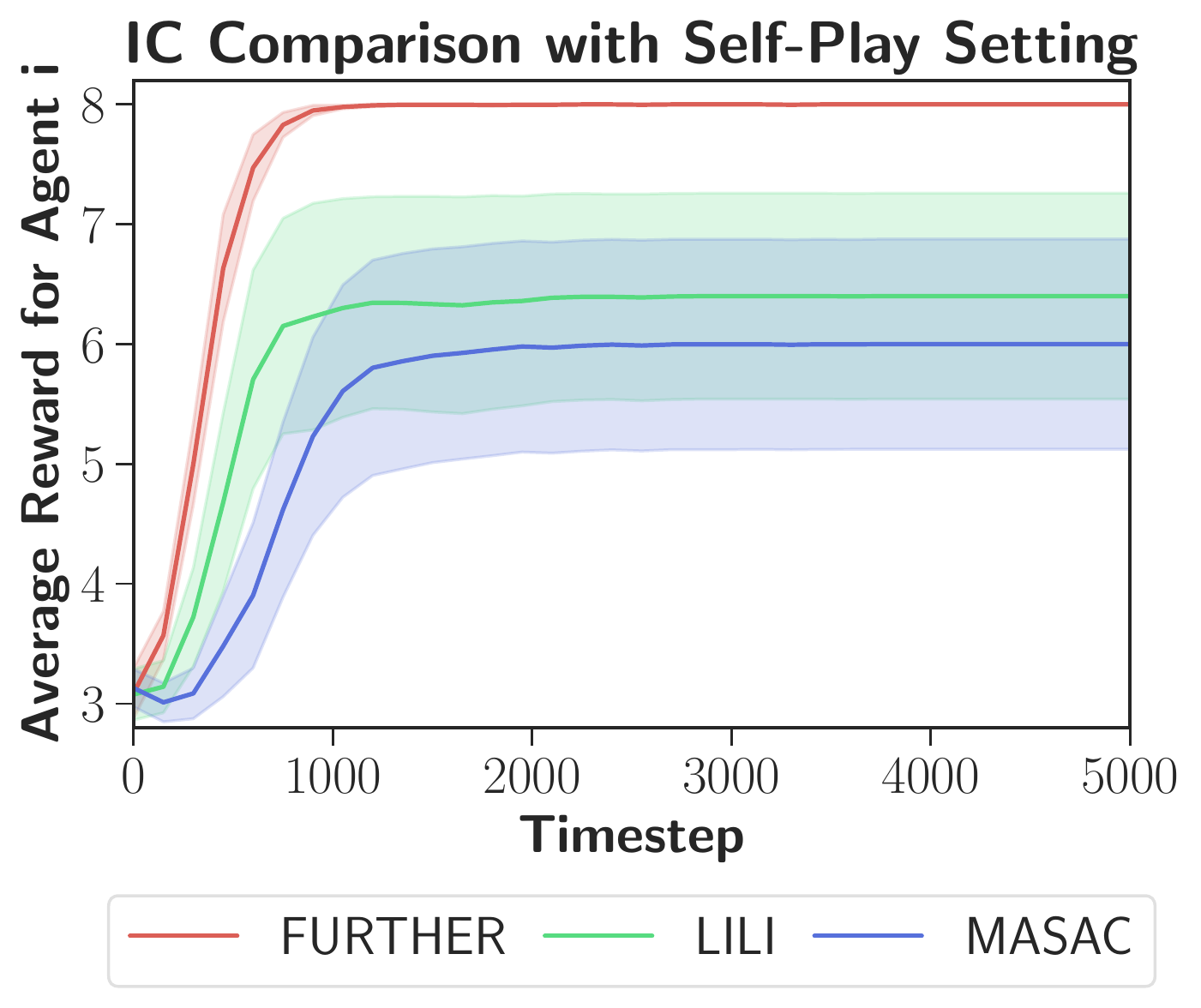}
        \caption{}
        \label{fig:ic-result}
    \end{subfigure}
    \begin{subfigure}[b]{0.33\linewidth}
        \centering
        \includegraphics[height=3.5cm]{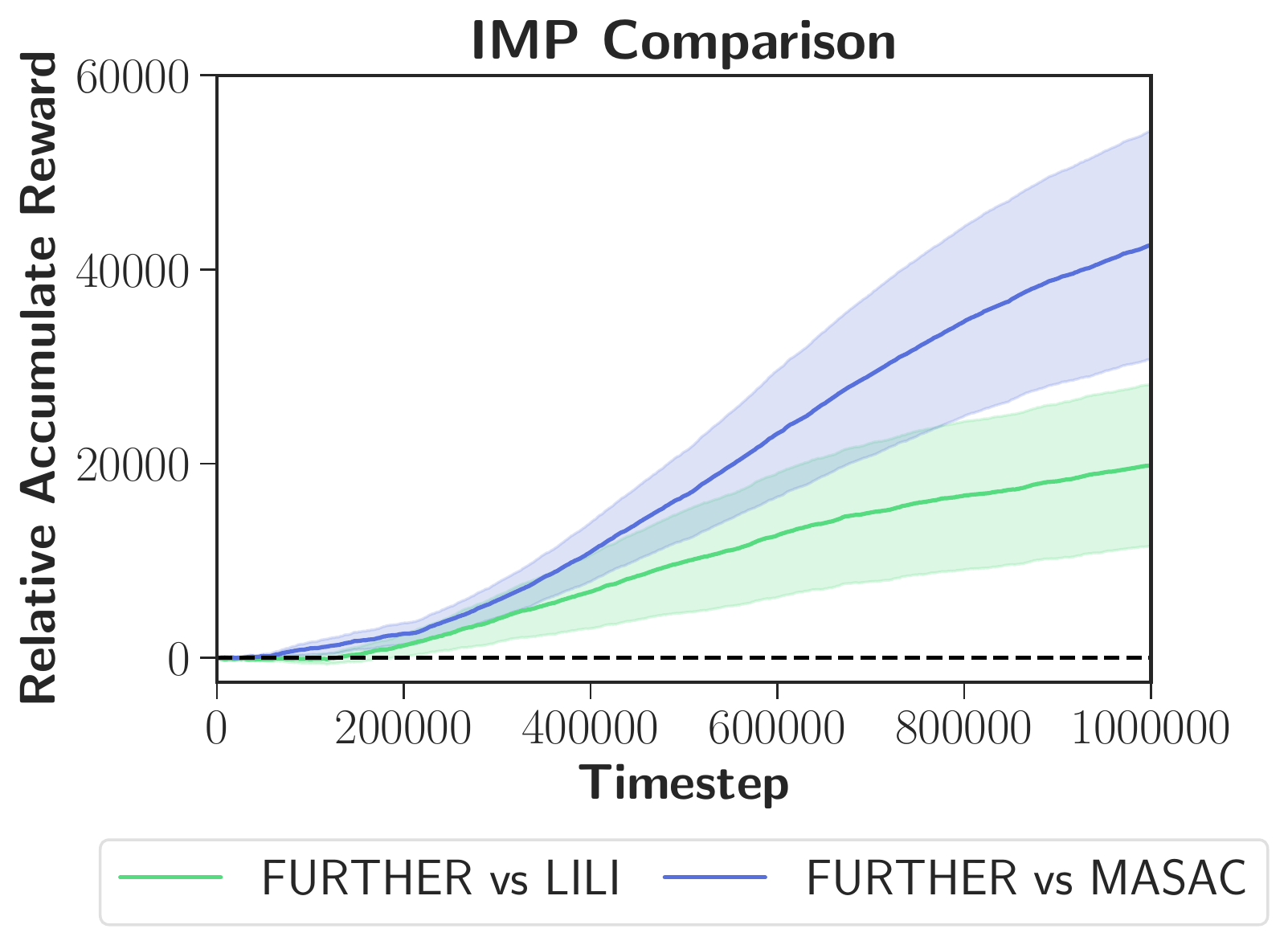}
        \caption{}
        \label{fig:imp-result}
    \end{subfigure}
    \vskip-0.09in
    \caption{\textbf{(a)} Convergence in IBS. The FURTHER agent achieves convergence to its optimal pure strategy Nash equilibrium. \textbf{(b)} Convergence in IC with self-play. The FURTHER team shows better converged performance than baselines. \textbf{(c)} A competitive play in IMP between FURTHER and baseline methods. FURTHER receives higher rewards than LILI and MASAC over time.}
    \vskip-0.15in
\end{figure}

\begin{figure}[t!]
\captionsetup[subfigure]{skip=0pt, aboveskip=3pt}
    \begin{subfigure}[b]{0.33\linewidth}
        \centering
        \includegraphics[height=3.35cm]{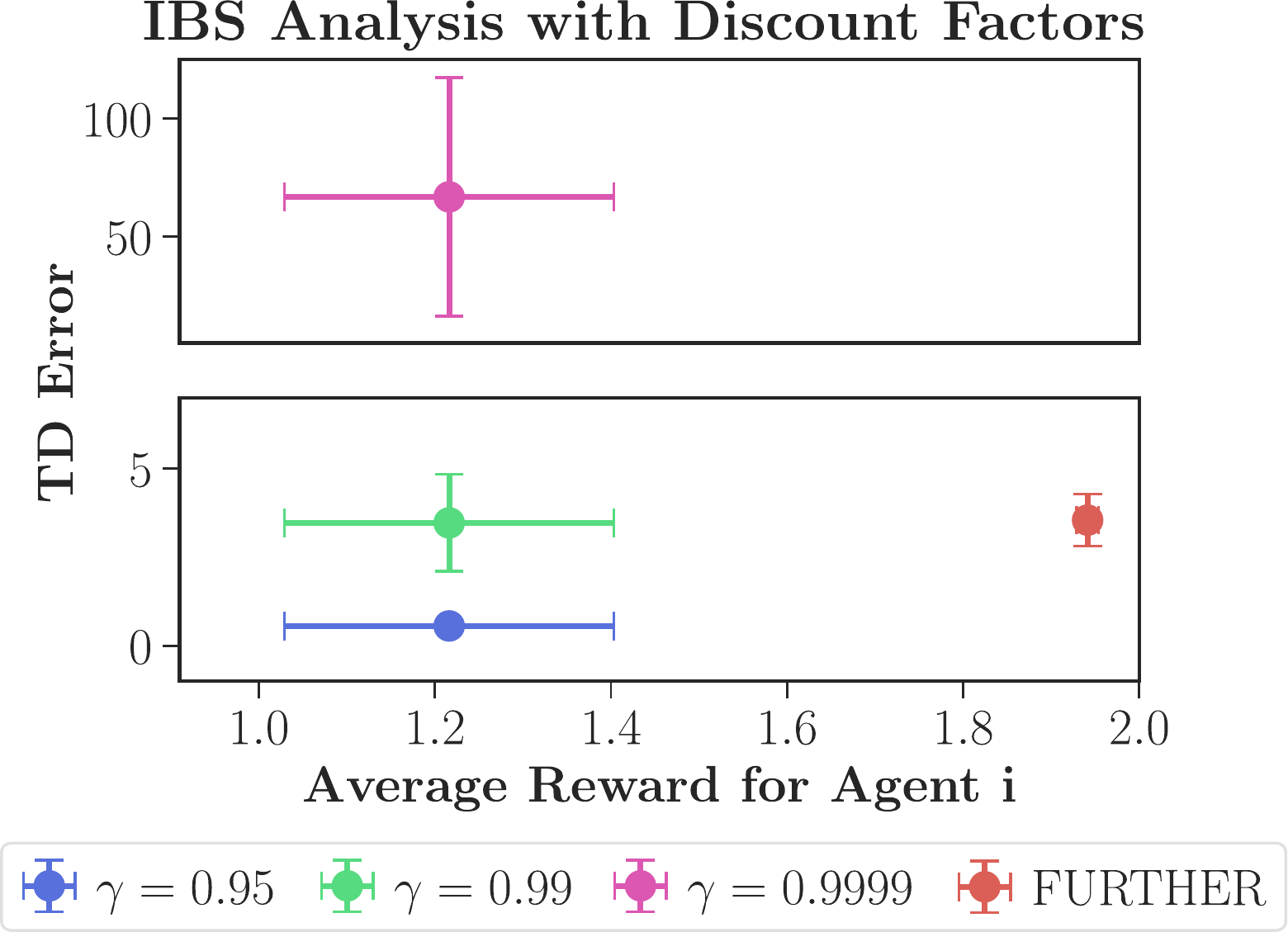}
        \caption{}
        \label{fig:discount-result}
    \end{subfigure}
    \begin{subfigure}[b]{0.33\linewidth}
        \centering
        \includegraphics[height=3.35cm]{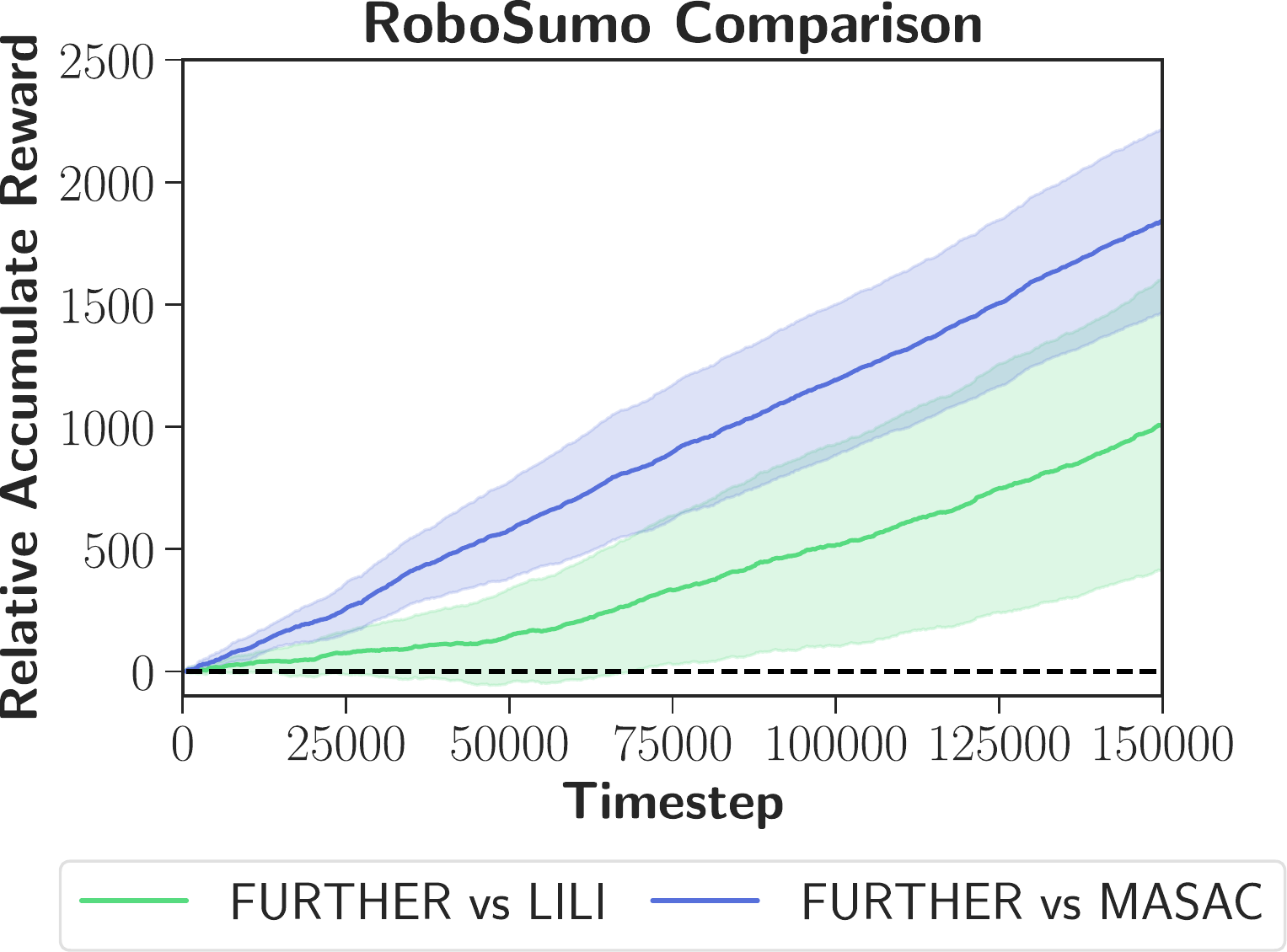}
        \caption{}
        \label{fig:robosumo-result}
    \end{subfigure}
    \begin{subfigure}[b]{0.33\linewidth}
        \centering
        \includegraphics[height=3.35cm]{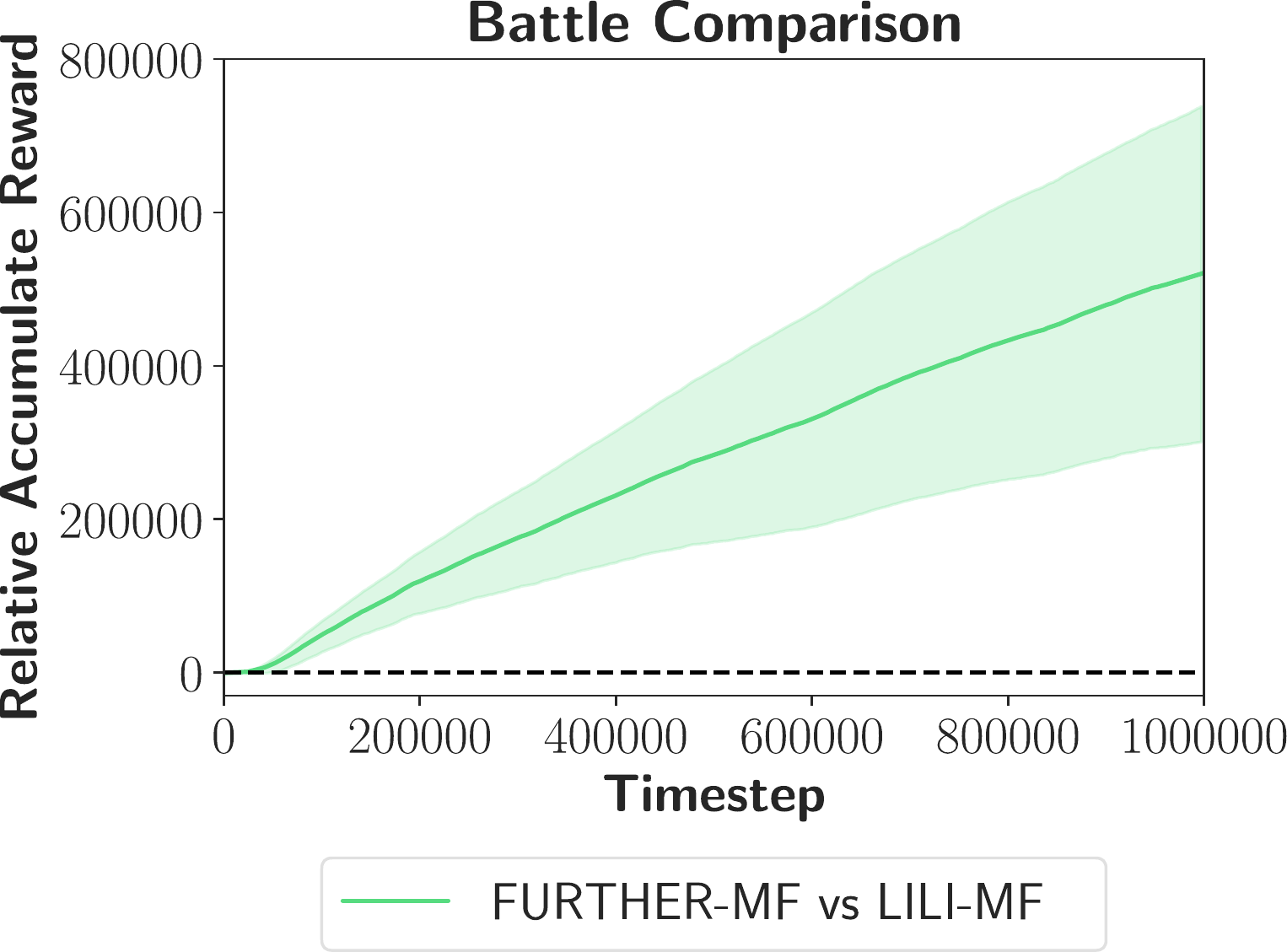}
        \caption{}
        \label{fig:battle-result}
    \end{subfigure}
    \vskip-0.1in
    \caption{\textbf{(a)} Convergence performance and corresponding TD errors with varying $\gamma$ in LILI. As $\gamma\!\rightarrow\!1$, LILI shows unstable learning (i.e., large TD error). \textbf{(b)} A competitive play in the RoboSumo domain, showing that FURTHER can learn a beneficial behavior in an environment with complex interactions \textbf{(c)} A mixed cooperative-competitive play in the battle domain. FURTHER-MF can solve a large-scale learning settings.}
    \vskip-0.2in
\end{figure}

The average reward performance when an agent $i$, trained with either FURTHER or the baseline methods, interacts with the $q$-learner $j$ is shown in~\Cref{fig:ibs-result}.
There are two notable observations. 
First, the FURTHER agent $i$ consistently converges to its optimal equilibrium of (B,B), while the LILI agent often converges to the sub-optimal equilibrium of (S,S). 
The FURTHER agent $i$ learns to select (B) while $j$ selects (S), receives the worst rewards of zero, and waits until $j$'s $q$-value for (S) is updated to be lower than the $q$-value for (B).
With the limiting view, $i$ learns that the waiting process is only temporary, and receiving the eventual rewards of 2 by converging to (B,B) is optimal. 
By contrast, LILI suffers from myopic evaluation and shows decreased performance upon convergence because the agent prefers simply converging to the sub-optimal equilibrium rather than waiting indefinitely. 
\Cref{fig:discount-result} also shows that LILI achieves sub-optimal performance for any value of $\gamma$ and shows unstable learning as $\gamma\!\rightarrow\!1$.
Second, FURTHER and LILI outperform the other approach of MASAC, showing the benefit of considering the active influence on future policies of other agents.

\paragraph{Question 2.} \textit{Which equilibrium do methods converge to in a self-play setting?}

We experiment with a self-play setting in which both agents learn with the same algorithm in an iterated cooperative (IC) game with identical payoffs (see~\Cref{fig:ic-domain}). 
This game has two pure strategy Nash equilibria of (U,U) and (D,D), in which the (D,D) equilibrium Pareto dominates the other.
\Cref{fig:ic-result} shows the average reward performance as the training iteration increases. 
First we find that LILI performs better than MASAC by considering the learning of agents. However, similar to the IBS results, we observe that FURTHER consistently converges to the best equilibrium of (D,D) while the baseline methods can converge to the sub-optimal equilibrium of (U,U) due to the myopic view. 

\paragraph{Question 3.} \textit{How does FURTHER's limiting optimization perform directly against baselines?}

We consider FURTHER agent $i$ directly competing against either LILI or MASAC opponent $j$ in the iterated matching pennies (IMP) game (see~\Cref{fig:imp-domain}). 
To show that FURTHER has a long-term perspective and thus can collect more rewards than the opposing method over time, we evaluate using a metric of relative accumulated reward summed up to the current timestep: $\smallsum\nolimits_{t}r^{i}_{t}\!-\!r^{j}_{t}$.
\Cref{fig:imp-result} shows that the relative accumulated reward for FURTHER is positive for both settings, meaning that FURTHER receives higher rewards than LILI and MASAC over time. 
This result suggests that FURTHER is more effective than LILI by employing the limiting view via the average reward formulation. 
This result also conveys that it is beneficial to consider the underlying learning dynamics rather than ignoring them because FURTHER can more easily exploit the MASAC opponent and achieve higher accumulated rewards than when competing against the LILI opponent.

\newpage
\paragraph{Question 4.} \textit{How does FURTHER scale to a more complex environment?}

To answer this question, we use the MuJoCo RoboSumo domain (\citep{alshedivat2018continuous}; see~\cref{fig:robosumo-domain}), where two ant robots compete with each other with the objective of pushing the opponent out of the ring. 
The reward function consists of a sparse reward of $5$ for winning against the opponent and shaped rewards of moving towards the opponent and pushing the opponent further from the center of the ring. 
This environment has complex interactions because an agent must learn how to control its joints with continuous action space to move around the ring while learning to push the opponent.
Similar to the setup in Question 3, FURTHER agent $i$ directly competes against either LILI or MASAC opponent $j$. 
We note that each agent has only partial observations about its opponent. As such, an agent infers its opponent's hidden policies and learning dynamics based on partial observations.
We show the RoboSumo results in~\Cref{fig:robosumo-result}. 
Consistent with our results in the iterated matrix games, we observe that FURTHER gains more rewards than the baselines over time and wins against MASAC more often than against LILI. 
The averaged winning rate across the entire interaction shows that FURTHER wins against LILI and MASAC with 60.6\% and 63.9\%, respectively. 
Therefore, FURTHER provides a scalable framework that can learn policies in an environment with complex interactions and continuous action space.

\paragraph{Question 5.} \textit{How does FURTHER scale to a large number of agents?}

Finally, we show the scalability of our method regarding the number of agents using the battle domain (\citep{zheng2018magent}; see~\cref{fig:battle-domain}).
In this large-scale mixed cooperative-competitive setting, a red team of 25 agents and a blue team of 25 agents interact in a gridworld, where each agent collaborates with its teammates to eliminate the opponents. 
Specifically, we compare when red and blue agents learn with the mean-field version of FURTHER (i.e., FURTHER-MF) and LILI (i.e., LILI-MF), respectively, where they predict the mean actions of neighboring agents.
We note that all 50 agents learn in a decentralized manner without sharing parameters with one another. 
It is evident that FURTHER-MF outperforms LILI-MF, which shows the effectiveness of having the limiting perspective. 
This result also conveys that FURTHER can easily incorporate other techniques in multiagent learning and show improved performance in large-scale settings. 

%% file: related_work.tex
\section{Related Work}
\paragraph{Stationary MARL.} 
The standard approach for addressing non-stationarity in MARL is to consider information about other agents and reason about joint action effects~\citep{hernandezLealK17survey}. 
Example frameworks include centralized training with decentralized execution, which accounts for the actions of other agents through a centralized critic~\citep{lowe17maddpg,forester17coma,iqbal19masac,yang18mean-field-marl,omidshafiei19teach,wadhwania2019policy,kim20hmat}.
Other related approaches include opponent modeling frameworks that infer opponent policies and condition an agent's policy on this inferred information about others~\citep{he16opponent-modeling,raileanu18opponent-modeling,grover18policy-representation,wen2018probabilistic}.
While this does alleviate non-stationarity, each agent learns its policy by assuming that other agents will follow the same policy into the future.
This assumption is incorrect because other agents can have different behavior in the future due to their learning~\citep{foerster17lola}, resulting in instability with respect to their changing behavior.
In contrast, FURTHER models the learning processes of other agents and considers how to actively influence limiting behavior.

\paragraph{Learning-aware MARL.} 
Our framework is closely related to prior work that considers the learning of other agents in the environment. 
The framework by~\cite{zhang10lookahead}, for instance, learns the best response adaptation to the other agent's anticipated updated policy. 
Notably, LOLA~\citep{foerster17lola} and its more recent improvements~\citep{letcher2018stable,foerster2018dice} study the impact of behavior on one or a few of another agent's policy updates.
Our work is also related to frameworks that leverage the inferred policy dynamics of other agents to impact their future policies by maximizing the discounted return objective~\citep{xie20lili,wang2021influencing}.
Meta-learning frameworks are also related that directly account for the non-stationary policy dynamics in multiagent settings based on the inner-loop and outer-loop optimization~\citep{kim21metamapg,lu2022modelfree,alshedivat2018continuous,balaguer2022good}.
Lastly, the field of incentive MARL~\cite{jaques19social,wang19,yang20,yang22} is related, where agents additionally optimize incentive rewards and learn successful policies in solving sequential social dilemma domains~\cite{leibo17socialdilemmas,wang19ssd}.
However, all of these approaches only account for a finite number of updates to the policies of other agents, so we observe that these methods can converge to a less desirable solution.
FURTHER addresses this issue by optimizing for the average reward objective in the active Markov game setting.

\paragraph{Game-theoretic MARL.} 
Another effective approach to addressing the non-stationarity is learning equilibrium policies that correspond to game-theoretic solution concepts~\citep{zinkevich06cyclic,littman94markov,littman01friendfoe,wang02nash,greenwald03correlated}. 
These frameworks predict stationary joint action values by the end of learning and can guarantee convergence to Nash~\citep{Nash48} or correlated~\citep{correlated87} equilibrium values under certain assumptions. 
However, as noted in~\cite{bowling05convergence}, this convergence is guaranteed only while ignoring the actual learning dynamics of other agents, and each agent assumes all agents will play the same joint equilibrium strategy. 
As such, equilibrium learners can fail to learn best-response policies when others choose to play different equilibrium strategies in the future as a result of their learning. 
By contrast, FURTHER considers convergence to a recurrent set of joint policies by inferring the true policy dynamics of other agents. 

%% file: conclusion.tex
\section{Conclusion}
In this paper, we have introduced FURTHER to address non-stationarity by considering each agent's impact on the converged policies of other agents.
The key idea is to consider the limiting policies of other agents through the average reward formulation for a newly proposed active Markov game framework, and we have developed a practical model-free and decentralized approach in this setting.
We evaluated our method on various multiagent settings and showed that FURTHER consistently converges to more desirable long-term behavior than state-of-the-art baseline approaches.

%% file: checklist.tex
\section*{Checklist}

\begin{enumerate}
\item For all authors...
\begin{enumerate}
  \item Do the main claims made in the abstract and introduction accurately reflect the paper's contributions and scope?
    \answerYes{}
  \item Did you describe the limitations of your work?
    \answerYes{See~\Cref{sec:limitation-and-impact}.}
  \item Did you discuss any potential negative societal impacts of your work?
    \answerYes{See~\Cref{sec:limitation-and-impact}.}
  \item Have you read the ethics review guidelines and ensured that your paper conforms to them?
    \answerYes{}
\end{enumerate}

\item If you are including theoretical results...
\begin{enumerate}
  \item Did you state the full set of assumptions of all theoretical results?
    \answerYes{}
        \item Did you include complete proofs of all theoretical results?
    \answerYes{See~\Cref{sec:proof-stochastically-stable-equilibrium,sec:bellman-proof,sec:policy-gradient-proof,sec:elbo-derivation}.}
\end{enumerate}

\item If you ran experiments...
\begin{enumerate}
  \item Did you include the code, data, and instructions needed to reproduce the main experimental results (either in the supplemental material or as a URL)?
    \answerYes{}
  \item Did you specify all the training details (e.g., data splits, hyperparameters, how they were chosen)?
    \answerYes{See~\Cref{sec:experiment-details}.}
        \item Did you report error bars (e.g., with respect to the random seed after running experiments multiple times)?
    \answerYes{The mean and 95\% confidence interval computed for 20 seeds are shown in each figure.}
        \item Did you include the total amount of compute and the type of resources used (e.g., type of GPUs, internal cluster, or cloud provider)?
    \answerYes{See~\Cref{sec:experiment-details}.}
\end{enumerate}

\item If you are using existing assets (e.g., code, data, models) or curating/releasing new assets...
\begin{enumerate}
  \item If your work uses existing assets, did you cite the creators?
    \answerYes{}
  \item Did you mention the license of the assets?
    \answerYes{}
  \item Did you include any new assets either in the supplemental material or as a URL?
    \answerYes{}
  \item Did you discuss whether and how consent was obtained from people whose data you're using/curating?
    \answerNA{}
  \item Did you discuss whether the data you are using/curating contains personally identifiable information or offensive content?
    \answerNA{}
\end{enumerate}

\item If you used crowdsourcing or conducted research with human subjects...
\begin{enumerate}
  \item Did you include the full text of instructions given to participants and screenshots, if applicable?
    \answerNA{}
  \item Did you describe any potential participant risks, with links to Institutional Review Board (IRB) approvals, if applicable?
    \answerNA{}
  \item Did you include the estimated hourly wage paid to participants and the total amount spent on participant compensation?
    \answerNA{}
\end{enumerate}
\end{enumerate}

%% file: appendix.tex
\appendix
\section{Example of Initial Condition Sensitivity}\label{sec:multichain-example}
\vskip-0.15in
\begin{figure}[h]
\captionsetup[subfigure]{skip=0pt, aboveskip=0pt}
    \begin{subfigure}[b]{0.49\linewidth}
        \centering
        \includegraphics[height=4.5cm]{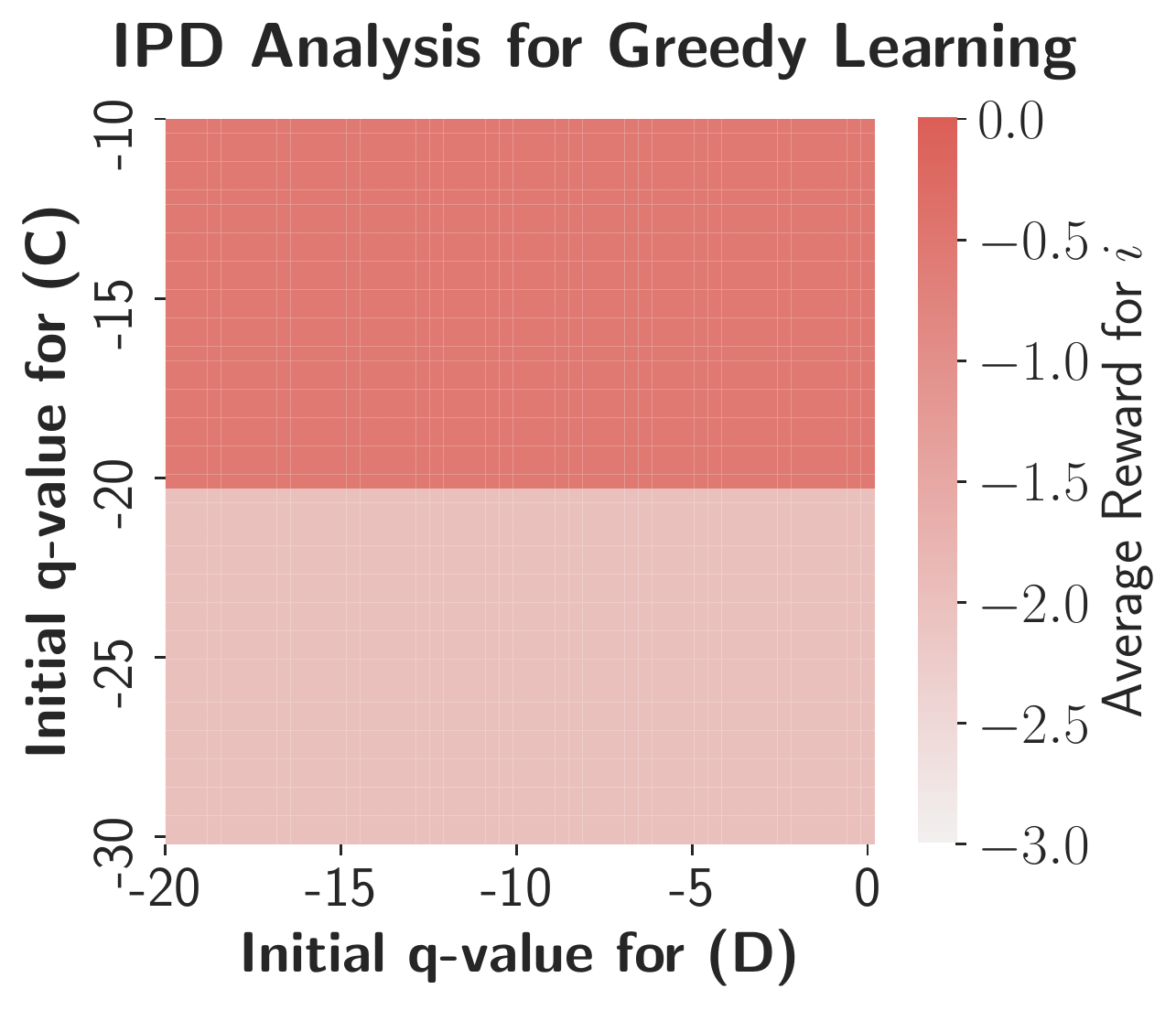}
        \caption{}
        \label{fig:policy-iteration-greedy}
    \end{subfigure}
    \begin{subfigure}[b]{0.49\linewidth}
        \centering
        \includegraphics[height=4.5cm]{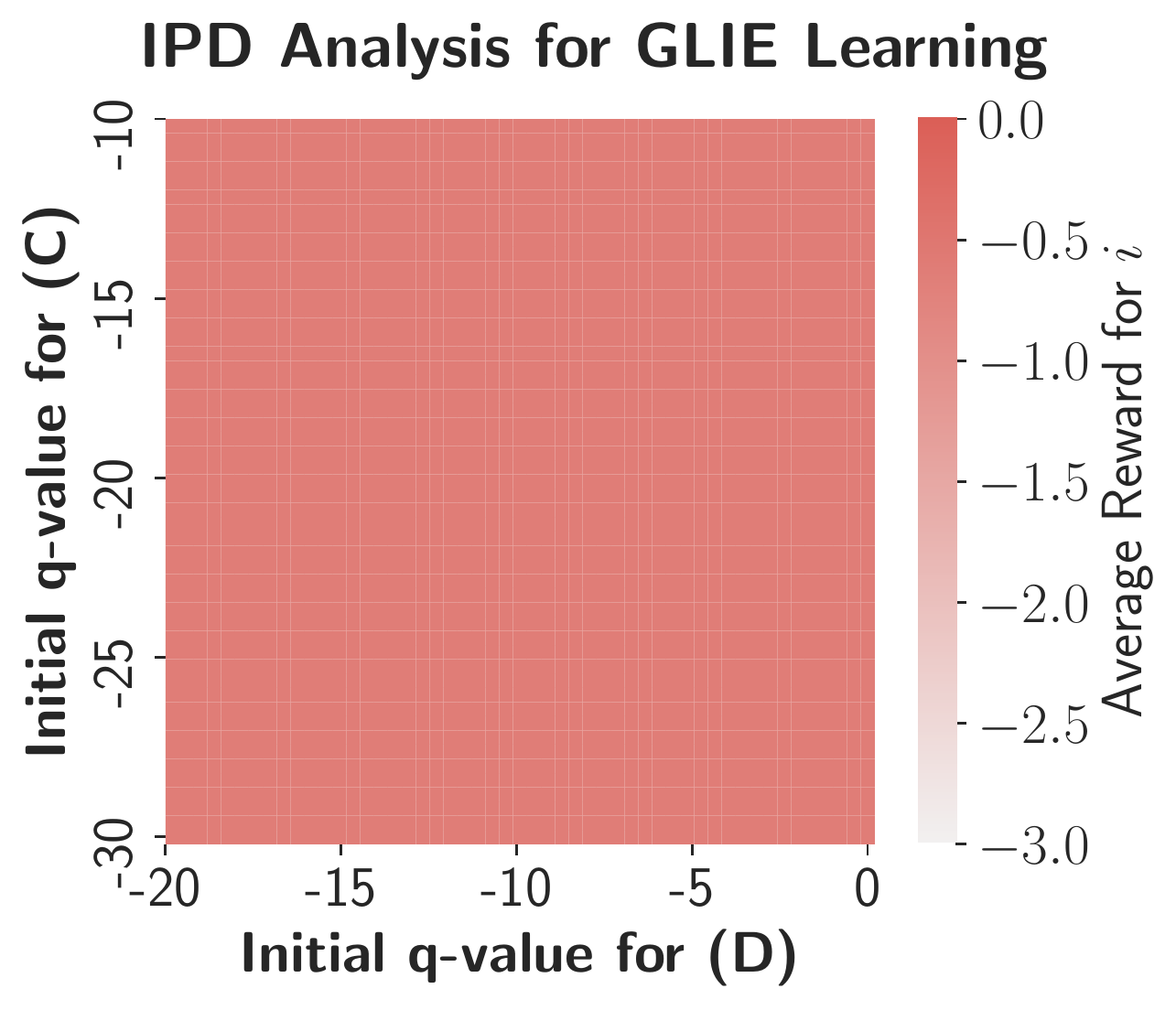}
        \caption{}
        \label{fig:policy-iteration-glie}
    \end{subfigure}
    \vskip-0.1in
    \caption{
    \textbf{(a)} A policy iteration analysis in IPD when agent $j$ has a greedy learning algorithm. Depending on $\bm{\theta^{\shortminus i}_{0}}$, $i$'s possible maximum average reward is affected. \textbf{(b)} A policy iteration analysis in IPD when agent $j$ has a GLIE learning algorithm. The possible maximum average reward for agent $i$ is independent to $j$'s initial policy $\bm{\theta^{\shortminus i}_{0}}$.}
    \vskip-0.1in
\end{figure}
Consider playing the iterated prisoner's dilemma (IPD) game (see~\Cref{tab:prisoner-dilemma-game}), where agent $i$ plays against a $q$-learning agent $j$. 
We perform a policy iteration analysis~\citep{puterman94mdp} with respect to
\begin{wraptable}[7]{r}{0.3\linewidth}
\vspace*{-0.12in}
\footnotesize
\centering
\begin{tabular}[b]{cc|cc}
\multicolumn{2}{c}{} & \multicolumn{2}{c}{}\\
\parbox[t]{2mm}{\multirow{3}{*}{}} 
    &       & $C$       & $D$       \\\cline{2-4}
\rule{0pt}{12pt}    & $C$   & $(\shortminus 1,\shortminus 1)$ & $(\shortminus 3, 0)$  \\
    & $D$   & $(0, \shortminus 3)$  & $(\shortminus 2,\shortminus 2)$ 
\end{tabular}
\vspace*{-0.13in}
\caption{Prisoner's dilemma game payoff matrix.}
\label{tab:prisoner-dilemma-game}
\end{wraptable}
 $j$'s varying initial $q$-values for each action $\bm{\theta^{\shortminus i}_{0}}$. 
\Cref{fig:policy-iteration-greedy} and \Cref{fig:policy-iteration-glie} show agent $i$'s average reward after convergence with respect to $\bm{\theta^{\shortminus i}_{0}}$ when $j$ trains with a greedy and GLIE algorithm, respectively.
Interestingly, the analysis with the greedy algorithm shows that $i$'s average reward depends on $\bm{\theta}^{\bm{\shortminus i}}_{\bm{0}}$ in IPD, where there is a set of $j$'s initial policies that $i$ can achieve the high average reward, but there is the other set of initial policies that can result in the undesirable average reward of $\shortminus 2$. 
By contrast, \Cref{fig:policy-iteration-glie} shows that $i$'s average reward is independent of $\bm{\theta}^{\bm{\shortminus i}}_{\bm{0}}$ when $j$'s learning satisfies GLIE, empirically supporting our discussion in \Cref{sec:stochastically-stable-distribution}.

\section{Uniqueness of Jointly-Stable Periodic Distribution}\label{sec:proof-stochastically-stable-equilibrium}
\textbf{Proposition 1.} (Uniqueness of Jointly-Stable Periodic Distribution). \textit{
Given communicating state transition $\mathcal{T}$ and perturbed joint update function with noise $\bm{\mathcal{U}_{\epsilon}}$, the jointly-stable periodic distribution is unique as $\epsilon\!\rightarrow\!0$ over time.}

\textit{Proof.} \hspace{0.2em} A perturbed Markov process has a unique stochastically stable distribution as noise $\epsilon\!\rightarrow\!0$ over time if a perturbed Markov process is regular: the transition matrix corresponding to a stationary policy contains a single recurrent class of states (i.e., states that are visited infinitely often) and a possibly empty set of transient states (i.e., states that are visited only finitely often) \cite{wicks05ssd} (Corollary 4.8, Section 5). 
As such, we prove that a Markov process of an active Markov game is regular by contradiction and thus show that the jointly-stable periodic distribution is unique as $\epsilon\!\rightarrow\!0$.
Suppose a perturbed Markov process of an active Markov game is irregular (i.e., there is more than one recurrent class), where the corresponding Markov matrix over the joint space of states and policies is defined as $p(s^{\prime},\bm{\theta^{\prime}}|s,\bm{\theta})\!=\!\smallsum_{\bm{a}}\bm{\pi}(\bm{a}|s;\bm{\theta})\mathcal{T}(s^{\prime}|s,\bm{a})\bm{\mathcal{U}_\epsilon}(\bm{\theta^{\prime}}|\bm{\theta},\bm{\tau})\;\forall s,s^{\prime}\!\in\!\mathcal{S},\bm{\theta},\bm{\theta^{\prime}}\!\in\!\bm{\Theta}$.
Because the perturbed joint update function has communicating strategies and thus contains a single recurrent class of policies, the state transition $\mathcal{T}$ must have multiple recurrent classes to result in an irregular active Markov game.
However, $\mathcal{T}$ has a single recurrent class only due to the communicating assumption, which is the contradiction. 
Therefore, we conclude that a perturbed Markov process of an active game is regular, which has a unique stochastically stable distribution as $\epsilon\!\rightarrow\!0$ by~\citep{wicks05ssd}.
\QEDB

\newpage
\section{Derivation of Active Differential Bellman Equation}\label{sec:bellman-proof}
\textbf{Proposition 2.} (Active Differential Bellman Equation). \textit{
The differential value function $v^{i}_{\theta^{i}}$ represents the expected total difference between the accumulated rewards from $s$ and $\bm{\theta}^{\bm{\shortminus i}}$ and the average reward $\rho^{i}_{\theta^{i}}$~\citep{sutton98rlbook}. The differential value function inherently includes the recursive relationship with respect to $v^{i}_{\theta^{i}}$ at the next state $s^{\prime}$ and the updated policies of other agents $\bm{\theta}^{\bm{\shortminus i\prime}}$}:
\begin{align*}
\begin{split}
v^{i}_{\theta^{i}}(s,\bm{\theta}^{\bm{\shortminus i}})&=
\lim_{T\rightarrow\infty}\mathbb{E}\Big[\smallsum_{t=0}^{T}\big(\mathcal{R}^{i}(s_t,\bm{a}_{\bm{t}})-\rho^{i}_{\theta^{i}}\big)\Big|\;\substack{s_{0}=s,\;\bm{\theta}^{\bm{\shortminus i}}_{\bm{0}}=\bm{\theta}^{\bm{\shortminus i}},\\a^{i}_{0:T}\sim\pi(\cdot|s_{0:T};\theta^{i}),\bm{a}^{\bm{\shortminus i}}_{\bm{0:T}}\sim\bm{\pi}(\bm{\cdot}|s_{0:T};\bm{\theta}^{\bm{\shortminus i}}_{\bm{0:T}}),\\s_{t+1}\sim\mathcal{T}(\cdot|s_{t},\bm{a_{t}}),\bm{\theta^{\shortminus i}_{t+1}}\sim\bm{\mathcal{U}^{\shortminus i}}(\cdot|\bm{\theta^{\shortminus i}_{t}},\bm{\tau^{\shortminus i}_{t}})}\;\Big]\\
&=\!\smallsum_{a^{i}}\pi(a^{i}|s;\theta^{i})\!\smallsum_{\bm{a}^{\bm{\shortminus i}}}\bm{\pi}(\bm{a}^{\bm{\shortminus i}}|s;\bm{\theta}^{\bm{\shortminus i}})\!\smallsum_{s'}\mathcal{T}(s'|s,\bm{a})\!\smallsum_{\bm{\theta}^{\bm{\shortminus i\prime}}}\bm{\mathcal{U}}{}^{\bm{\shortminus i}}(\bm{\theta}^{\bm{\shortminus i\prime}}|\bm{\theta}^{\bm{\shortminus i}},\bm{\tau^{\shortminus i}})\\& 
\quad\Big[
\mathcal{R}^{i}(s,\bm{a})-\rho^{i}_{\theta^{i}}+v^{i}_{\theta^{i}}(s',\bm{\theta}^{\bm{\shortminus i\prime}})\Big].
\end{split}
\end{align*}

\textit{Proof.} \hspace{0.2em}We seek to derive the recursive relationship between $v^{i}_{\theta^{i}}(s,\bm{\theta}^{\bm{\shortminus i}})$ and $v^{i}_{\theta^{i}}(s^{\prime},\bm{\theta}^{\bm{\shortminus i\prime}})$. 
We leverage the general derivation outlined in~\cite{sutton98rlbook} (page 59) and extend it to our active Markov game formulation:
\begin{align}
\begin{split}
v^{i}_{\theta^{i}}(s,\bm{\theta}^{\bm{\shortminus i}})\!&=\!\lim_{T\rightarrow\infty}\!\mathbb{E}\Big[\smallsum_{t=0}^{T}\big(\mathcal{R}^{i}(s_t,\bm{a}_{\bm{t}})-\rho^{i}_{\theta^{i}}\big)\Big|\;\substack{s_{0}=s,\;\bm{\theta}^{\bm{\shortminus i}}_{\bm{0}}=\bm{\theta}^{\bm{\shortminus i}},\\a^{i}_{0:T}\sim\pi(\cdot|s_{0:T};\theta^{i}),\bm{a}^{\bm{\shortminus i}}_{\bm{0:T}}\sim\bm{\pi}(\bm{\cdot}|s_{0:T};\bm{\theta}^{\bm{\shortminus i}}_{\bm{0:T}}),\\s_{t+1}\sim\mathcal{T}(\cdot|s_{t},\bm{a_{t}}),\bm{\theta^{\shortminus i}_{t+1}}\sim\bm{\mathcal{U}^{\shortminus i}}(\cdot|\bm{\theta^{\shortminus i}_{t}},\bm{\tau^{\shortminus i}_{t}})}\;\Big]\\
&=\lim_{T\rightarrow\infty}\!\mathbb{E}\Big[\mathcal{R}^{i}(s_0,\bm{a_\bm{0}})\!-\!\rho^{i}_{\theta^{i}}+\smallsum_{t=1}^{T}\big(\mathcal{R}^{i}(s_t,\bm{a^{\shortminus i}_t})-\rho^{i}_{\theta^{i}}\big)\Big|\;\substack{s_{0}=s,\;\bm{\theta}^{\bm{\shortminus i}}_{\bm{0}}=\bm{\theta}^{\bm{\shortminus i}},\\a^{i}_{0:T}\sim\pi(\cdot|s_{0:T};\theta^{i}),\bm{a}^{\bm{\shortminus i}}_{\bm{0:T}}\sim\bm{\pi}(\bm{\cdot}|s_{0:T};\bm{\theta}^{\bm{\shortminus i}}_{\bm{0:T}}),\\s_{t+1}\sim\mathcal{T}(\cdot|s_{t},\bm{a_{t}}),\bm{\theta^{\shortminus i}_{t+1}}\sim\bm{\mathcal{U}^{\shortminus i}}(\cdot|\bm{\theta^{\shortminus i}_{t}},\bm{\tau^{\shortminus i}_{t}})}\Big]\\
&=\smallsum_{a^{i}}\pi(a^{i}|s;\theta^{i})\smallsum_{\bm{a}^{\bm{\shortminus i}}}\bm{\pi}(\bm{a}^{\bm{\shortminus i}}|s;\bm{\theta}^{\bm{\shortminus i}})\smallsum_{s'}\mathcal{T}(s'|s,\bm{a})\smallsum_{\bm{\theta}^{\bm{\shortminus i\prime}}}\bm{\mathcal{U}}^{\bm{\shortminus i}}(\bm{\theta}^{\bm{\shortminus i\prime}}|\bm{\theta}^{\bm{\shortminus i}},\bm{\tau}^{\bm{\shortminus i}})\\
&\quad\Big[\mathcal{R}^{i}(s,\bm{a})\!-\!\rho^{i}_{\theta^{i}}\!+\!\!\lim_{T\rightarrow\infty}\!\!\mathbb{E}\Big[\!\smallsum_{t=0}^{T}\!\big(\mathcal{R}^{i}(s_{t+1},\bm{a}_{\bm{t+1}})\!-\!\rho^{i}_{\theta^{i}}\big)\Big|\;\substack{s_{1}=s^{\prime},\;\bm{\theta}^{\bm{\shortminus i}}_{\bm{1}}=\bm{\theta}^{\bm{\shortminus i\prime}},\\a^{i}_{1:T}\sim\pi(\cdot|s_{1:T};\theta^{i}),\bm{a}^{\bm{\shortminus i}}_{\bm{1:T}}\sim\bm{\pi}(\bm{\cdot}|s_{1:T};\bm{\theta}^{\bm{\shortminus i}}_{\bm{1:T}}),\\s_{t+1}\sim\mathcal{T}(\cdot|s_{t},\bm{a_{t}}),\bm{\theta^{\shortminus i}_{t+1}}\sim\bm{\mathcal{U}^{\shortminus i}}(\cdot|\bm{\theta^{\shortminus i}_{t}},\bm{\tau^{\shortminus i}_{t}})}\Big]\Big]\\
&=\smallsum_{a^{i}}\pi(a^{i}|s;\theta^{i})\smallsum_{\bm{a}^{\bm{\shortminus i}}}\bm{\pi}(\bm{a}^{\bm{\shortminus i}}|s;\bm{\theta}^{\bm{\shortminus i}})\smallsum_{s'}\mathcal{T}(s'|s,\bm{a})\smallsum_{\bm{\theta}^{\bm{\shortminus i\prime}}}\bm{\mathcal{U}}{}^{\bm{\shortminus i}}(\bm{\theta}^{\bm{\shortminus i\prime}}|\bm{\theta}^{\bm{\shortminus i}},\bm{\tau}^{\bm{\shortminus i}})\\& 
\quad\Big[
\mathcal{R}^{i}(s,\bm{a})\!-\!\rho^{i}_{\theta^{i}}\!+\!v^{i}_{\theta^{i}}(s',\bm{\theta}^{\bm{\shortminus i\prime}})\Big].
\raisetag{17pt}
\end{split}
\end{align}
\QEDB
\section{Derivation of Active Average Reward Policy Gradient}\label{sec:policy-gradient-proof}
\textbf{Proposition 3.} (Active Average Reward Policy Gradient Theorem). \textit{
The gradient of active average reward objective in~\Cref{eqn:further-objective} with respect to agent $i$'s policy parameters $\theta^i$ is}:
\begin{gather*}
\nabla_{\theta^{i}}J^{i}_{\pi}(\theta^{i})\!=\!\frac{1}{k}\!\smallsum_{\ell=1}^{k}\smallsum_{s_{\ell},\bm{\theta^{\shortminus i}_{\ell}}}\!\mu_{k,\theta^{i}}(s_{\ell},\bm{\theta_{\ell}}|\ell)\!\smallsum_{a^{i}_{\ell}}\nabla_{\theta^{i}}\pi(a^{i}_{\ell}|s_{\ell};\theta^{i})\!\smallsum_{\bm{a_{\ell}}^{\bm{\shortminus i}}}\bm{\pi}(\bm{a_{\ell}}^{\bm{\shortminus i}}|s_{\ell};\bm{\theta_{\ell}}^{\bm{\shortminus i}})q^{i}_{\theta^{i}}(s_{\ell},\bm{\theta_{\ell}}^{\bm{\shortminus i}},\bm{a_{\ell}}),\\
\text{with}\quad\!\!\!\! q^{i}_{\theta^{i}}(s_{\ell},\bm{\theta_{\ell}}^{\bm{\shortminus i}},\bm{a_{\ell}})\!=\!\!\smallsum_{s_{\ell+1}}\!\!\mathcal{T}(s_{\ell+1}|s_{\ell},\bm{a_{\ell}})\!\!\smallsum_{\bm{\theta_{\ell+1}}^{\bm{i}}}\!\!\bm{\mathcal{U}^{\shortminus i}}(\bm{\theta_{\ell+1}}^{\bm{\shortminus i}}|\bm{\theta_{\ell}}^{\bm{\shortminus i}},\bm{\tau_{\ell}}^{\bm{\shortminus i}})\Big[\mathcal{R}^{i}(s_{\ell},\bm{a_{\ell}})\!-\!\rho^{i}_{\theta^{i}}\!+\!v^{i}_{\theta^{i}}(s_{\ell+1},\bm{\theta_{\ell+1}}^{\bm{\shortminus i}})\Big].
\end{gather*}

\textit{Proof.} \hspace{0.2em}We seek to derive an expression for optimizing the average reward objective in~\Cref{eqn:further-objective} with respect to agent $i$'s policy parameters $\theta^{i}$. 
Our derivation leverages the general policy gradient theorem proof for the continuing case in \cite{sutton98rlbook} (page 334).
We begin by expressing the gradient of the differential value function $v^{i}_{\theta^{i}}(s,\bm{\theta}^{\bm{\shortminus i}})$ for $s\!\in\!\mathcal{S}$ and $\bm{\theta^{\shortminus i}}\!\in\!\bm{\Theta^{\shortminus i}}$:
\begin{align}\label{eqn:policy-gradient-step1}
\begin{split}
\nabla_{\theta^{i}}v^{i}_{\theta^{i}}(s,\bm{\theta}^{\bm{\shortminus i}})&=\nabla_{\theta^{i}}\Big[\smallsum_{a^{i}}\pi(a^{i}|s;\theta^{i})\smallsum_{\bm{a}^{\bm{\shortminus i}}}\bm{\pi}(\bm{a}^{\bm{\shortminus i}}|s;\bm{\theta}^{\bm{\shortminus i}})q^{i}_{\theta^{i}}(s,\bm{\theta}^{\bm{\shortminus i}},\bm{a})\Big]\\
&=\smallsum_{a^{i}}\nabla_{\theta^{i}}\pi(a^{i}|s;\theta^{i})\smallsum_{\bm{a}^{\bm{\shortminus i}}}\bm{\pi}(\bm{a}^{\bm{\shortminus i}}|s;\bm{\theta}^{\bm{\shortminus i}})q^{i}_{\theta^{i}}(s,\bm{\theta}^{\bm{\shortminus i}},\bm{a})+\\
&\quad\,\smallsum_{a^{i}}\pi(a^{i}|s;\theta^{i})\smallsum_{\bm{a}^{\bm{\shortminus i}}}\bm{\pi}(\bm{a}^{\bm{\shortminus i}}|s;\bm{\theta}^{\bm{\shortminus i}})\underbrace{\nabla_{\theta^{i}}q^{i}_{\theta^{i}}(s,\bm{\theta}^{\bm{\shortminus i}},\bm{a})}_{\text{Term A}}.
\end{split}
\end{align}
We continue to derive the Term A in \Cref{eqn:policy-gradient-step1}:
\begin{align}\label{eqn:policy-gradient-step2}
\begin{split}
\nabla_{\theta^{i}}q^{i}_{\theta^{i}}(s,\bm{\theta}^{\bm{\shortminus i}},\bm{a})\!&=\!\nabla_{\theta^{i}}\!\Big[\smallsum_{s'}\mathcal{T}(s'|s,\bm{a})\smallsum_{\bm{\theta}^{\bm{\shortminus i\prime}}}\bm{\mathcal{U}}{}^{\bm{\shortminus i}}(\bm{\theta}^{\bm{\shortminus i\prime}}|\bm{\theta}^{\bm{\shortminus i}},\bm{\tau}^{\bm{\shortminus i}})\Big[\mathcal{R}^{i}(s,\bm{a})\!-\!\rho^{i}_{\theta^{i}}\!+\!v^{i}_{\theta^{i}}(s^{\prime},\bm{\theta}^{\bm{\shortminus i\prime}})\Big]\Big]\\
&=-\nabla_{\theta^{i}}\rho^{i}_{\theta^{i}}+\smallsum_{s'}\!\mathcal{T}(s'|s,\bm{a})\!\smallsum_{\bm{\theta}^{\bm{\shortminus i\prime}}}\bm{\mathcal{U}}^{\bm{\shortminus i}}(\bm{\theta}^{\bm{\shortminus i\prime}}|\bm{\theta}^{\bm{\shortminus i}},\bm{\tau}^{\bm{\shortminus i}})\nabla_{\theta^{i}}v^{i}_{\theta^{i}}(s^{\prime},\bm{\theta}^{\bm{\shortminus i\prime}}).
\raisetag{20pt}
\end{split}
\end{align}
We summarize~\Cref{eqn:policy-gradient-step1} and \Cref{eqn:policy-gradient-step2} together and re-arrange terms to obtain:
\begin{align}\label{eqn:policy-gradient-step3}
\begin{split}
\nabla_{\theta^{i}}\rho^{i}_{\theta^{i}}
&=\smallsum_{a^{i}}\nabla_{\theta^{i}}\pi(a^{i}|s;\theta^{i})\smallsum_{\bm{a}^{\bm{\shortminus i}}}\bm{\pi}(\bm{a}^{\bm{\shortminus i}}|s;\bm{\theta}^{\bm{\shortminus i}})q^{i}_{\theta^{i}}(s,\bm{\theta}^{\bm{\shortminus i}},\bm{a})+\\
&\quad\,\smallsum_{a^{i}}\pi(\cdot|s;\theta^{i})\smallsum_{\bm{a}^{\bm{\shortminus i}}}\bm{\pi}(\bm{a}^{\bm{\shortminus i}}|s;\bm{\theta}^{\bm{\shortminus i}})\smallsum_{s'}\mathcal{T}(s'|s,\bm{a})\smallsum_{\bm{\theta}^{\bm{\shortminus i\prime}}}\bm{\mathcal{U}}{}^{\bm{\shortminus i}}(\bm{\theta}^{\bm{\shortminus i\prime}}|\bm{\theta}^{\bm{\shortminus i}},\bm{\tau}^{\bm{\shortminus i}})\nabla_{\theta^{i}}v^{i}_{\theta^{i}}(s^{\prime},\bm{\theta}^{\bm{\shortminus i\prime}})-\\
&\quad\,\,\nabla_{\theta^{i}}v^{i}_{\theta^{i}}(s,\bm{\theta}^{\bm{\shortminus i}}).
\raisetag{13pt}
\end{split}
\end{align}
We define the jointly-stable periodic distribution with respect to the agent $i$'s fixed stationary policy: 
\begin{align}\label{eqn:jointly-stable-periodic-distribution-wrt-fixed-i}
\begin{split}
\frac{1}{k}\smallsum_{\ell=1}^{k}&\mu_{k,\theta^{i}}(s_{\ell+1},\bm{\theta}_{\bm{\ell+1}}|\ell\!+\!1)\!=\!\frac{1}{k}\smallsum_{\ell=1}^{k}\smallsum_{s_{\ell},\bm{\theta^{\shortminus i}_{\ell}}}\mu_{k,\theta^{i}}(s_{\ell},\bm{\theta_{\ell}}|\ell)\smallsum_{\bm{a}_{\bm{\ell}}}\bm{\pi}(\bm{a_\ell}|s_{\ell};\bm{\theta_{\ell}})\\
&\;\mathcal{T}(s_{\ell+1}|s_{\ell},\bm{a_{\ell}})\;\bm{\mathcal{U}^{\shortminus i}}(\bm{\theta^{\shortminus i}_{\ell+1}}|\bm{\theta^{\shortminus i}_{\ell}},\bm{\tau^{\shortminus i}_{\ell}})\quad\forall s_{\ell+1}\!\in\!\mathcal{S},\bm{\theta_{\ell+1}}\!\in\!\bm{\Theta},
\end{split}
\end{align}
where $\bm{\theta_{\ell}}\!=\!\{\theta^{i},\bm{\theta^{\shortminus i}_{\ell}}\}$.
We now apply \Cref{eqn:jointly-stable-periodic-distribution-wrt-fixed-i} to \Cref{eqn:policy-gradient-step3} and derive the final expression for policy gradient by writing $\nabla_{\theta^{i}}\rho^{i}_{\theta^{i}}$ as $\nabla_{\theta^{i}}J^{i}_{\pi}(\theta^{i})$:
\begin{align}\label{eqn:policy-gradient-step4}
\begin{split}
&\frac{1}{k}\smallsum_{\ell=1}^{k}\smallsum_{s_{\ell},\bm{\theta^{\shortminus i}_{\ell}}}\mu_{k,\theta^{i}}(s_{\ell},\bm{\theta_{\ell}}|\ell)\nabla_{\theta^{i}}J^{i}_{\pi}(\theta^{i})\!=\!\cfrac{1}{k}\smallsum_{\ell=1}^{k}\smallsum_{s_{\ell},\bm{\theta^{\shortminus i}_{\ell}}}\mu_{k,\theta^{i}}(s_{\ell},\bm{\theta_{\ell}}|\ell)\Big[\\
&\quad\smallsum_{a^{i}_{\ell}}\nabla_{\theta^{i}}\pi(a^{i}_{\ell}|s_{\ell};\theta^{i})\smallsum_{\bm{a_{\ell}}^{\bm{\shortminus i}}}\bm{\pi}(\bm{a_{\ell}}^{\bm{\shortminus i}}|s_{\ell};\bm{\theta_{\ell}}^{\bm{\shortminus i}})q^{i}_{\theta^{i}}(s_{\ell},\bm{\theta_{\ell}}^{\bm{\shortminus i}},\bm{a_{\ell}})+\\
&\quad\smallsum_{a^{i}_{\ell}}\pi(a^{i}_{\ell}|s_{\ell};\theta^{i})\smallsum_{\bm{a_{\ell}}^{\bm{\shortminus i}}}\bm{\pi}(\bm{a_{\ell}}^{\bm{\shortminus i}}|s_{\ell};\bm{\theta_{\ell}}^{\bm{\shortminus i}})\smallsum_{s_{\ell+1}}\!\!\mathcal{T}(s_{\ell+1}|s_{\ell},\bm{a_{\ell}})\!\!\smallsum_{\bm{\theta_{\ell+1}}^{\bm{\shortminus i}}}\!\!\bm{\mathcal{U}}^{\bm{\shortminus i}}(\bm{\theta_{\ell+1}}^{\bm{\shortminus i}}|\bm{\theta_{\ell}}^{\bm{\shortminus i}},\bm{\tau_{\ell}}^{\bm{\shortminus i}})\nabla_{\theta^{i}}v^{i}_{\theta^{i}}(s_{\ell+1},\bm{\theta_{\ell+1}}^{\bm{\shortminus i}})-\\
&\quad\;\nabla_{\theta^{i}}v^{i}_{\theta^{i}}(s_{\ell},\bm{\theta_{\ell}}^{\bm{\shortminus i}})\Big].
\raisetag{20pt}
\end{split}
\end{align}
Note that the left-hand side $\nabla_{\theta^{i}}J^{i}_{\pi}(\theta^{i})$ does not depend on $s_{\ell}$ and $\bm{\theta^{\shortminus i}_{\ell}}$, so \Cref{eqn:policy-gradient-step4} becomes:
\begin{align}
\begin{split}
&\nabla_{\theta^{i}}J^{i}_{\pi}(\theta^{i})\\
&=\frac{1}{k}\smallsum_{\ell=1}^{k}\smallsum_{s_{\ell},\bm{\theta^{\shortminus i}_{\ell}}}\mu_{k,\theta^{i}}(s_{\ell},\bm{\theta_{\ell}}|\ell)\smallsum_{a^{i}_{\ell}}\nabla_{\theta^{i}}\pi(a^{i}_{\ell}|s_{\ell};\theta^{i})\smallsum_{\bm{a_{\ell}}^{\bm{\shortminus i}}}\bm{\pi}(\bm{a_{\ell}}^{\bm{\shortminus i}}|s_{\ell};\bm{\theta_{\ell}}^{\bm{\shortminus i}})q^{i}_{\theta^{i}}(s_{\ell},\bm{\theta_{\ell}}^{\bm{\shortminus i}},\bm{a_{\ell}})+\\
&\quad\,\,\frac{1}{k}\smallsum_{\ell=1}^{k}\smallsum_{s_{\ell},\bm{\theta^{\shortminus i}_{\ell}}}\mu_{k,\theta^{i}}(s_{\ell},\bm{\theta_{\ell}}|\ell)\smallsum_{a^{i}_{\ell}}\pi(a^{i}_{\ell}|s_{\ell};\theta^{i})\smallsum_{\bm{a_{\ell}}^{\bm{\shortminus i}}}\bm{\pi}(\bm{a_{\ell}}^{\bm{\shortminus i}}|s_{\ell};\bm{\theta_{\ell}}^{\bm{\shortminus i}})\smallsum_{s_{\ell+1}}\mathcal{T}(s_{\ell+1}|s_{\ell},\bm{a_{\ell}})\\
&\quad\quad\smallsum_{\bm{\theta_{\ell+1}}^{\bm{\shortminus i}}}\bm{\mathcal{U}}^{\bm{\shortminus i}}(\bm{\theta_{\ell+1}}^{\bm{\shortminus i}}|\bm{\theta_{\ell}}^{\bm{\shortminus i}},\bm{\tau_{\ell}}^{\bm{\shortminus i}})\nabla_{\theta^{i}}v^{i}_{\theta^{i}}(s_{\ell+1},\bm{\theta_{\ell+1}}^{\bm{\shortminus i}})-\\
&\quad\,\,\frac{1}{k}\smallsum_{\ell=1}^{k}\smallsum_{s_{\ell},\bm{\theta^{\shortminus i}_{\ell}}}\mu_{k,\theta^{i}}(s_{\ell},\bm{\theta_{\ell}}|\ell)\nabla_{\theta^{i}}v^{i}_{\theta^{i}}(s_{\ell},\bm{\theta_{\ell}}^{\bm{\shortminus i}})\\
&=\frac{1}{k}\smallsum_{\ell=1}^{k}\smallsum_{s_{\ell},\bm{\theta^{\shortminus i}_{\ell}}}\mu_{k,\theta^{i}}(s_{\ell},\bm{\theta_{\ell}}|\ell)\smallsum_{a^{i}_{\ell}}\nabla_{\theta^{i}}\pi(a^{i}_{\ell}|s_{\ell};\theta^{i})\smallsum_{\bm{a_{\ell}}^{\bm{\shortminus i}}}\bm{\pi}(\bm{a_{\ell}}^{\bm{\shortminus i}}|s_{\ell};\bm{\theta_{\ell}}^{\bm{\shortminus i}})q^{i}_{\theta^{i}}(s_{\ell},\bm{\theta_{\ell}}^{\bm{\shortminus i}},\bm{a_{\ell}})+\\
&\quad\,\,\frac{1}{k}\smallsum_{\ell=1}^{k}\smallsum_{\substack{s_{\ell+1},\\\bm{\theta^{\shortminus i}_{\ell+1}}}}\!\mu_{k,\theta^{i}}(s_{\ell+1},\bm{\theta_{\ell+1}}|\ell+1)\nabla_{\theta^{i}}v^{i}_{\theta^{i}}(s_{\ell+1},\bm{\theta_{\ell+1}}^{\bm{\shortminus i}})\!-\!\cfrac{1}{k}\smallsum_{\ell=1}^{k}\smallsum_{s_{\ell},\bm{\theta^{\shortminus i}_{\ell}}}\!\mu_{k,\theta^{i}}(s_{\ell},\bm{\theta_{\ell}}|\ell)\nabla_{\theta^{i}}v^{i}_{\theta^{i}}(s_{\ell},\bm{\theta_{\ell}}^{\bm{\shortminus i}})\\
&=\frac{1}{k}\smallsum_{\ell=1}^{k}\smallsum_{s_{\ell},\bm{\theta^{\shortminus i}_{\ell}}}\mu_{k,\theta^{i}}(s_{\ell},\bm{\theta_{\ell}}|\ell)\smallsum_{a^{i}_{\ell}}\nabla_{\theta^{i}}\pi(a^{i}_{\ell}|s_{\ell};\theta^{i})\smallsum_{\bm{a_{\ell}}^{\bm{\shortminus i}}}\bm{\pi}(\bm{a_{\ell}}^{\bm{\shortminus i}}|s_{\ell};\bm{\theta_{\ell}}^{\bm{\shortminus i}})q^{i}_{\theta^{i}}(s_{\ell},\bm{\theta_{\ell}}^{\bm{\shortminus i}},\bm{a_{\ell}}).
\raisetag{21pt}
\end{split}
\end{align}
\QEDB
\section{Additional Implementation Details}\label{sec:implementation-details}
\subsection{Network Structure}
Our neural networks for the inference learning and reinforcement learning module consist of fully-connected layers for vector observations (e.g., iterated matrix games, MuJoCo RoboSumo~\citep{alshedivat2018continuous}) and additional convolution layers for image observations (e.g., MAgent Battle~\cite{zheng2018magent}). 
The encoder outputs the mean and standard deviation for the Gaussian distribution of $p(\bm{\hat{z}}^{\bm{\shortminus i}}_{\bm{t+1}}|\bm{\hat{z}}^{\bm{\shortminus i}}_{\bm{t}},\tau^{i}_{t};\phi^{i}_{\text{enc}})$, where we sample $\bm{\hat{z}}^{\bm{\shortminus i}}_{\bm{t}}$ by applying the reparameterization trick~\citep{blei17variational}. 
From the sampled $\bm{\hat{z}}^{\bm{\shortminus i}}_{\bm{t}}$, the decoder 
$p(\bm{a}^{\bm{\shortminus i}}_{\bm{t}}|s_{t},\bm{\hat{z}}^{\bm{\shortminus i}}_{\bm{t}};\phi^{i}_{\text{dec}})$ outputs a probability for the categorical distribution (discrete action space) or a mean and variance for the Gaussian distribution (continuous action space).
Similarly, the policy $\pi(a^{i}_{t}|s_{t},\bm{\hat{z}_{t}}^{\bm{\shortminus i}};\theta^{i})$ outputs a probability for the categorical distribution (discrete action space) or a mean and variance for the Gaussian distribution (continuous action space).
Lastly, the critic outputs $q$-values for all actions for discrete action space (i.e., $q^{i}_{\theta^{i}}(a^{i}_{t}|s_{t},\bm{\hat{z}_{t}}^{\bm{\shortminus i}},\bm{a^{\shortminus i}_{t}};\psi^{i}_{\beta})$) by following~\cite{christodoulou2019soft} or outputs a $q$-value given the joint action for continuous action space (i.e., $q^{i}_{\theta^{i}}(s_{t},\bm{\hat{z}_{t}}^{\bm{\shortminus i}},\bm{a_{t}};\psi^{i}_{\beta})$). 

\subsection{Optimization} 
We detail additional notes about our implementation:
\begin{itemize}[leftmargin=*, wide, labelindent=0pt, topsep=0pt]
    \itemsep 0in 
    \item For simplicity, we consider the period $k\!=\!1$ and develop corresponding soft reinforcement learning optimizations in~\Cref{sec:sac-with-vi}. The current FURTHER implementation can be extended to settings with $k\!>\!1$ by sampling $k$ states and policies that are consecutive within each batch.
    \item For continuous action space, we modify SAC for continuous action space~\cite{haarnoja18sac} and replace the soft value function $v^{i}_{\theta^{i}}$ in~\Cref{eqn:soft-v-value-optimization} with:
    \begin{align}\label{eqn:soft-v-value-optimization-continuous}
    \begin{split}
    v^{i}_{\theta^{i}}\!(s,\bm{\hat{z}}^{\bm{\shortminus i}};\psi^{i})\!=\!\!\mathbb{E}_{a^{i}\sim\pi(\cdot|s,\bm{\hat{z}}^{\bm{\shortminus i}};\theta^{i}),\bm{a}^{\bm{\shortminus i}}\sim\pi(\bm{\cdot}|s;\bm{\hat{z}}^{\bm{\shortminus i}})}\!\big[\!\min\limits_{\beta=1,2}\!q^{i}_{\theta^{i}}\!(s,\bm{\hat{z}}^{\bm{\shortminus i}}\!,\bm{a};\psi^{i}_{\beta})\big]\!+\!\alpha\mathcal{H}(\pi(\cdot|s,\bm{\hat{z}}^{\bm{\shortminus i}};\theta^{i})).
    \raisetag{20pt}
    \end{split}
    \end{align}
    We also replace the policy optimization in \Cref{eqn:soft-policy-optimization} with the following:
    \begin{align}\label{eqn:soft-policy-optimization-continuous}
    \begin{split}
    J^{i}_{\pi}(\theta^{i})\!&=\!\mathbb{E}_{(s,\bm{\hat{z}}^{\bm{\shortminus i}},\bm{a}^{\bm{\shortminus i}})\sim\mathcal{D}^{i},\epsilon\sim\mathcal{N}(0,I)}\!\Big[\\
    &\min\limits_{\beta=1,2}q^{i}_{\theta^{i}}(s,\bm{\hat{z}}^{\bm{\shortminus i}},f_{\theta^{i}}(\epsilon; s, \bm{\hat{z}}^{\bm{\shortminus i}}),\bm{a^{\shortminus i}};\psi^{i}_{\beta})\!-\!\alpha\log\pi(f_{\theta^{i}}(\epsilon; s, \bm{\hat{z}}^{\bm{\shortminus i}})|s,\bm{\hat{z}}^{\bm{\shortminus i}};\theta^{i})\Big],
    \end{split}
    \end{align}
    where $a^{i}\!=\!f_{\theta^{i}}(\epsilon; s, \bm{\hat{z}}^{\bm{\shortminus i}})$ denotes the output of the reparameterized $i$'s policy \cite{haarnoja18sac}.
    \item In practice, we apply a weighting of $0.01$ on the KL divergence term in~\Cref{eqn:elbo} for balanced training of the inference learning module.
    \item Because it is impractical to consider the entire interactions from the beginning of the game in computing~\Cref{eqn:elbo}, we limit $\tau^{i}_{0:t-1}$ to be recent interactions specified by a batch size.
\end{itemize}
\newpage

\subsection{Pseudocode} 
\begin{algorithm}[H]
	\caption{FURTHER and FURTHER Mean-Field}\label{alg:algorithm}  
	\small
	\begin{algorithmic}[1]
	    \REQUIRE Learning rates $\alpha_q,\alpha_\rho,\alpha_{\pi},\alpha_{\phi}$, soft $q$-target update rate $\tau_{q}$
	    \STATE \textit{\# Agent initialization}
		\FOR{Each agent $i$}
	        \State Initialize RL module $\theta^{i},\psi^{i}_{1},\psi^{i}_{2},\bar{\psi}^{i}_{1},\bar{\psi}^{i}_{2},\rho^{i}_{\theta^i},\mathcal{D}^{i}$
	        \State Initialize inference module $\phi^{i}_{\text{enc}},\phi^{i}_{\text{dec}}$
	        \State Initialize other agents' latent strategies $\bm{\hat{z}}^{\bm{\shortminus i}}_{\bm{0}}$
		\ENDFOR
		\FOR{Each timestep $t$}
	        \STATE \textit{\# Decentralized execution}
		    \FOR{Each agent $i$}
		        \STATE Select action $a^{i}_{t}\sim\pi(\cdot|s_{t},\bm{\hat{z}}^{\bm{\shortminus i}}_{\bm{t}};\theta^{i})$ 
		    \ENDFOR
		    \STATE Execute joint action $\bm{a_{t}}$ and receive next state $s_{t+1}$ and joint rewards $\bm{r_{t}}$
		    \STATE \textit{\# Mean action computation and perform inference}
		    \FOR{Each agent $i$}
		        \IF{Apply mean-field}
		            \STATE Compute mean action of its neighborhood $\bar{a}_{t}^{\shortminus i}$ and set $\bm{a_t}\!=\!\{a^{i}_{t},\bar{a}_{t}^{\shortminus i}\}$ 
		        \ENDIF
		        \STATE Infer next updated policies of other agents $\bm{\hat{z}}^{\bm{\shortminus i}}_{\bm{t+1}}\sim p(\cdot|\bm{\hat{z}}^{\bm{\shortminus i}}_{\bm{t}},\tau^{i}_{t};\phi^{i}_{\text{enc}})$
		        \State Add a transition to its replay memory $\mathcal{D}^{i}\!\leftarrow\!\mathcal{D}^{i}\!\cup\!\{s_{t},\bm{\hat{z}}^{\bm{\shortminus i}}_{\bm{t}},\bm{a_{t}},r^{i}_{t},s_{t+1},\bm{\hat{z}}^{\bm{\shortminus i}}_{\bm{t+1}}\}$ 
		    \ENDFOR
	        \STATE \textit{\# Decentralized training}
		    \FOR{Each agent $i$}
		        \STATE $\{\psi^{i}_{\beta},\rho^{i}_{\theta^{i}}\}\leftarrow\{\psi^{i}_{\beta},\rho^{i}_{\theta^{i}}\}-\{\alpha_{q},\alpha_{\rho}\}J^{i}_{q}(\psi^{i}_{\beta},\rho^{i}_{\theta^{i}})$ for $\beta=1,2$
		        \STATE $\theta^{i}\leftarrow\theta^{i}+\alpha_{\pi}J^{i}_{\pi}(\theta^{i})$
		        \STATE $\{\phi^{i}_{\text{enc}},\phi^{i}_{\text{dec}}\}\leftarrow\{\phi^{i}_{\text{enc}},\phi^{i}_{\text{dec}}\}-\alpha_{\phi}J^{i}_{\text{elbo}}(\phi^{i}_{\text{enc}},\phi^{i}_{\text{dec}})$
		        \STATE $\bar{\psi}^{i}_{\beta}\leftarrow \tau_{q}\psi^{i}_{\beta}+(1-\tau_{q})\bar{\psi}^{i}_{\beta}$ for $\beta=1,2$
		    \ENDFOR
		\ENDFOR
	\end{algorithmic}
\end{algorithm}

\section{ELBO Derivation}\label{sec:elbo-derivation}
We derive our ELBO optimization in~\Cref{eqn:elbo} for the inference module. In particular, we follow the ELBO derivation in \cite{Zintgraf2020VariBAD} (Appendix A) and modify it for our multiagent setting:

\begin{align}\label{eqn:elbo-derivation}
\begin{split}
\mathbb{E}_{p(\tau^{i}_{0:t})}\Big[\log p(\tau^{i}_{1:t};\phi^{i}_{\text{dec}})\Big]&=\mathbb{E}_{p(\tau^{i}_{0:t})}\Big[\log\int p(\tau^{i}_{1:t},\bm{\hat{z}}^{\bm{\shortminus i}}_{\bm{1:t}};\phi^{i}_{\text{dec}})d\bm{\hat{z}}^{\bm{\shortminus i}}_{\bm{1:t}}\Big]\\
&=\mathbb{E}_{p(\tau^{i}_{0:t})}\Big[\log\int p(\tau^{i}_{1:t},\bm{\hat{z}}^{\bm{\shortminus i}}_{\bm{1:t}};\phi^{i}_{\text{dec}})\frac{p(\bm{\hat{z}}^{\bm{\shortminus i}}_{\bm{1:t}}|\tau^{i}_{0:t\!-\!1};\phi^{i}_{\text{enc}})}{p(\bm{\hat{z}}^{\bm{\shortminus i}}_{\bm{1:t}}|\tau^{i}_{0:t\!-\!1};\phi^{i}_{\text{enc}})}d\bm{\hat{z}}^{\bm{\shortminus i}}_{\bm{1:t}}\Big]\\
&=\mathbb{E}_{p(\tau^{i}_{0:t})}\Big[\log\mathbb{E}_{p(\bm{\hat{z}}^{\bm{\shortminus i}}_{\bm{1:t}}|\tau^{i}_{0:t\!-\!1};\phi^{i}_{\text{enc}})}\big[\frac{p(\tau^{i}_{1:t},\bm{\hat{z}}^{\bm{\shortminus i}}_{\bm{1:t}};\phi^{i}_{\text{dec}})}{p(\bm{\hat{z}}^{\bm{\shortminus i}}_{\bm{1:t}}|\tau^{i}_{0:t\!-\!1};\phi^{i}_{\text{enc}})}\big]\Big]\\
&\geq\mathbb{E}_{p(\tau^{i}_{0:t}),p(\bm{\hat{z}}^{\bm{\shortminus i}}_{\bm{1:t}}|\tau^{i}_{0:t\!-\!1};\phi^{i}_{\text{enc}})}\Big[\log\frac{p(\tau^{i}_{1:t},\bm{\hat{z}}^{\bm{\shortminus i}}_{\bm{1:t}};\phi^{i}_{\text{dec}})}{p(\bm{\hat{z}}^{\bm{\shortminus i}}_{\bm{1:t}}|\tau^{i}_{0:t\!-\!1};\phi^{i}_{\text{enc}})}\Big]\\
&=\mathbb{E}_{p(\tau^{i}_{0:t}),p(\bm{\hat{z}}^{\bm{\shortminus i}}_{\bm{1:t}}|\tau^{i}_{0:t\!-\!1};\phi^{i}_{\text{enc}})}\Big[\log p(\tau^{i}_{1:t},\bm{\hat{z}}^{\bm{\shortminus i}}_{\bm{1:t}};\phi^{i}_{\text{dec}})-\log p(\bm{\hat{z}}^{\bm{\shortminus i}}_{\bm{1:t}}|\tau^{i}_{0:t\!-\!1};\phi^{i}_{\text{enc}})\Big]\\
&=\mathbb{E}_{p(\tau^{i}_{0:t}),p(\bm{\hat{z}}^{\bm{\shortminus i}}_{\bm{1:t}}|\tau^{i}_{0:t\!-\!1};\phi^{i}_{\text{enc}})}\Big[\smallsum_{t^{\prime}=1}^{t}\log p(\bm{a^{\shortminus i}_{t^{\prime}}}|s_{t^{\prime}},\bm{\hat{z}}^{\bm{\shortminus i}}_{\bm{t^{\prime}}};\phi^{i}_{\text{dec}})+\smallsum_{t^{\prime}=0}^{t-1}\log p(\bm{\hat{z}}^{\bm{\shortminus i}}_{\bm{t^{\prime}}})-\\
&\quad\quad\quad\quad\quad\quad\quad\quad\quad\quad\,\,\smallsum_{t^{\prime}=1}^{t}\log p(\bm{\hat{z}}^{\bm{\shortminus i}}_{\bm{t^{\prime}}}|\tau^{i}_{t^{\prime}\!-\!1};\phi^{i}_{\text{enc}})\Big].
\raisetag{20pt}
\end{split}
\end{align}
Finally, we summarize terms to obtain:
\begin{align*}
\begin{split}
\mathbb{E}_{p(\tau^{i}_{0:t}),p(\bm{\hat{z}}^{\bm{\shortminus i}}_{\bm{1:t}}|\tau^{i}_{0:t\!-\!1};\phi^{i}_{\text{enc}})}\Big[\smallsum_{t^{\prime}=1}^{t}\underbrace{\log p(\bm{a}^{\bm{\shortminus i}}_{\bm{t^{\prime}}}|s_{t^{\prime}},\bm{\hat{z}}^{\bm{\shortminus i}}_{\bm{t^{\prime}}};\phi^{i}_{\text{dec}})}_{\text{Reconstruction loss}}-\underbrace{D_{\text{KL}}\big(p(\bm{\hat{z}}^{\bm{\shortminus i}}_{\bm{t^{\prime}}}|\tau^{i}_{t^{\prime}\!-\!1};\phi^{i}_{\text{enc}})||p(\bm{\hat{z}}^{\bm{\shortminus i}}_{\bm{t^{\prime}\!-\!1}})\big)}_{\text{KL divergence}}\Big].
\end{split}
\end{align*}

\section{Experimental and Hyperparameter Details}\label{sec:experiment-details}
\subsection{Domain Details}
\paragraph{Iterated matrix games.} As in~\cite{foerster17lola}, we model the state space in all iterated matrix games as $s_{0}\!=\!\varnothing$ and $s_{t}\!=\!\bm{a}_{\bm{t-1}}$ for $t\!\geq\!1$. 
For these simple domains, we empirically observe that training the policy and critics based on the most recent transition improves training performance.
Lastly, in Question 1, we consider agent $i$ playing against a $q$-learning agent $j$ with a learning rate $\alpha_q$ of $0.5$, a discount factor $\gamma$ of $0.9$, and a fixed $\epsilon$-exploration of $0.05$.

\paragraph{MuJoco RoboSumo.} Each ant robot observes a vector with size $128$, which consists of the position of its own and the opponent's body, its own joint angles and velocities, and forces exerted on each part of its own body and the opponent’s torso~\citep{alshedivat2018continuous}.
We note that each agent has partial observations about its opponent and cannot observe the opponent's velocities and limb positions. 
Regarding the action space, each agent has a continuous action space with a dimension of $8$. 
Lastly, we use the reward function that consists of a sparse reward of $5$ for winning against the opponent and the following shaped rewards:
\begin{itemize}[leftmargin=*, wide, labelindent=0pt, topsep=0pt]
    \item Reward for moving towards the opponent proportional to $\shortminus d_{\text{opp}}$, where $d_{\text{opp}}$ denotes the distance between the agent and the opponent.
    \item Reward for pushing the opponent further from the center of the ring proportional to $\text{exp}(\shortminus d_{\text{center}})$, where $d_{\text{center}}$ denotes the distance of the opponent from the center of the ring.
\end{itemize}
We refer to~\citep{alshedivat2018continuous} (Appendix D) for more RoboSumo details.

\paragraph{MAgent Battle.}
Each agent receives an observation of a $13\!\times\!13\!\times\!9$ image with the following channels: its and opponent's team presence, its and opponent's team HP, its and opponent's team minimap, and its position~\cite{zheng2018magent}. 
The discrete action space has a dimension of $21$ for moving around the gridworld and attacking the opponents.
Lastly, reward is given as $5$ for killing an opponent, $\shortminus 0.005$ for every timestep cost, $0,2$ for attacking an opponent, and $\shortminus 0.1$ reward for dying. 
We refer to~\citep{zheng2018magent} for more MAgent details.

\subsection{Baseline Details}\label{appendix:baseline-details}
\begin{itemize}[leftmargin=*, wide, labelindent=0pt, topsep=0pt]
    \item LILI~\citep{xie20lili} maximizes the discounted return $v^{i}_{\theta^{i}}$ in the active Markov game:
    \begin{align}\label{eqn:lili-objective}
    \begin{split}
    \max_{\theta^{i}}v^{i}_{\theta^i}(s,\bm{\theta}^{\bm{\shortminus i}})\!:=\!\max_{\theta^{i}}\mathbb{E}\Big[\smallsum_{t=0}^{\infty}\gamma^{t}\mathcal{R}^{i}(s_t,\bm{a}_{\bm{t}})\Big|\substack{s_{0}=s,\;\bm{\theta}^{\bm{\shortminus i}}_{\bm{0}}=\bm{\theta}^{\bm{\shortminus i}},\\a^{i}_{0:T}\sim\pi(\cdot|s_{0:T};\theta^{i}),\bm{a}^{\bm{\shortminus i}}_{\bm{0:T}}\sim\bm{\pi}(\cdot|s_{0:T};\bm{\theta}^{\bm{\shortminus i}}_{\bm{0:T}}),\\s_{t+1}\sim\mathcal{T}(\cdot|s_{t},\bm{a_{t}}),\bm{\theta^{\shortminus i}_{t+1}}\sim\bm{\mathcal{U}^{\shortminus i}}(\cdot|\bm{\theta^{\shortminus i}_{t}},\bm{\tau^{\shortminus i}_{t}})}\Big].
    \end{split}
    \end{align}
    We implement LILI by replacing the average reward target $y$ in \Cref{eqn:soft-q-value-optimization} with the discounted return target: $y\!=\!r^{i}+\gamma v^{i}_{\theta^{i}}(s',\bm{\hat{z}}^{\bm{\shortminus i\prime}};\bar{\psi}^{i}_{\beta})$.
    \item MASAC~\citep{iqbal19masac} maximizes the discounted return $v^{i}_{\theta^{i}}$ in the stationary Markov game:
    \begin{align}\label{eqn:masac-objective}
    \begin{split}
    \max_{\theta^{i}}\rho^{i}_{\theta^i}(s,\bm{\theta}^{\bm{\shortminus i}})\!:=\!\max_{\theta^{i}}\mathbb{E}\Big[\smallsum_{t=0}^{\infty}\gamma^{t}\mathcal{R}^{i}(s_t,\bm{a}_{\bm{t}})\Big|\substack{s_{0}=s,\\a^{i}_{0:T}\sim\pi(\cdot|s_{0:T};\theta^{i}),\bm{a}^{\bm{\shortminus i}}_{\bm{0:T}}\sim\bm{\pi}(\cdot|s_{0:T};\bm{\theta}^{\bm{\shortminus i}}),\\s_{t+1}\sim\mathcal{T}(\cdot|s_{t},\bm{a_{t}})}\Big].
    \end{split}
    \end{align}
    MASAC employs the framework of centralized training with decentralized execution~\citep{lowe17maddpg} and has access to other agents' policies to perform optimization during training.
    \item \textbf{DRON~\citep{he16dron}:} An approach that extends DQN~\citep{mnih13dqn} with opponent modeling by predicting both Q-values and current strategies of other agents. This baseline fails to predict future policies of others.
    \item \textbf{MOA~\citep{jaques19social}:} An approach that additionally optimizes the influence reward to consider influential actions to other agents. This baseline also has the discounted return objective.
\end{itemize}

\subsection{Hyperparameter Details}
We use an internal cluster equipped with GPUs of RTX 3090 and CPUs of AMD Threadripper 3960X for choosing hyperparameters. 
We report the important hyperparameter values that we used for each of the methods in our experiments:
\begin{table}[H]
\centering
\begin{tabular}{l|l}
Hyperparameter & Value \\ \hline
Critic learning rate $\alpha_{q}$ & 0.002\\
Gain learning rate $\alpha_{\rho}$ & 0.02\\
Actor learning rate $\alpha_{\pi}$ & 0.0005\\
Inference learning rate $\alpha_{\phi}$ & 0.002\\
Entropy weight $\alpha$ & 0.4\\
Dimension of latent space $|\bm{z^{\shortminus i}}|$ & 5 \\
Discount factor $\gamma$ & 0.99 \\
Batch size & 256 \\
\end{tabular}
\caption{IBS Experiment}
\end{table}

\begin{table}[H]
\centering
\begin{tabular}{l|l}
Hyperparameter & Value \\ \hline
Critic learning rate $\alpha_{q}$ & 0.0005\\
Gain learning rate $\alpha_{\rho}$ & 0.02\\
Actor learning rate $\alpha_{\pi}$ & 0.0001\\
Inference learning rate $\alpha_{\phi}$ & 0.0005\\
Entropy weight $\alpha$ & 0.3\\
Dimension of latent space $|\bm{z^{\shortminus i}}|$ & 5 \\
Discount factor $\gamma$ & 0.99 \\
Batch size & 64 \\
\end{tabular}
\caption{IC Experiment}
\end{table}

\begin{table}[H]
\centering
\begin{tabular}{l|l}
Hyperparameter & Value \\ \hline
Critic learning rate $\alpha_{q}$ & 0.01\\
Gain learning rate $\alpha_{\rho}$ & 0.05\\
Actor learning rate $\alpha_{\pi}$ & 0.001\\
Inference learning rate $\alpha_{\phi}$ & 0.01\\
Entropy weight $\alpha$ & 0.35\\
Dimension of latent space $|\bm{z^{\shortminus i}}|$ & 5 \\
Discount factor $\gamma$ & 0.99 \\
Batch size & 64 \\
\end{tabular}
\caption{IMP Experiment}
\end{table}

\begin{table}[H]
\centering
\begin{tabular}{l|l}
Hyperparameter & Value \\ \hline
Critic learning rate $\alpha_{q}$ & 0.0002\\
Gain learning rate $\alpha_{\rho}$ & 0.2\\
Actor learning rate $\alpha_{\pi}$ & 0.0001\\
Inference learning rate $\alpha_{\phi}$ & 0.0002\\
Entropy weight $\alpha$ & 0.01\\
Dimension of latent space $|\bm{z^{\shortminus i}}|$ & 10 \\
Discount factor $\gamma$ & 0.99 \\
Batch size & 256 \\
\end{tabular}
\caption{RoboSumo Experiment}
\end{table}

\begin{table}[H]
\centering
\begin{tabular}{l|l}
Hyperparameter & Value \\ \hline
Critic learning rate $\alpha_{q}$ & 0.001\\
Gain learning rate $\alpha_{\rho}$ & 0.2\\
Actor learning rate $\alpha_{\pi}$ & 0.0005\\
Inference learning rate $\alpha_{\phi}$ & 0.001\\
Entropy weight $\alpha$ & 0.01\\
Dimension of latent space $|\bm{z^{\shortminus i}}|$ & 10 \\
Discount factor $\gamma$ & 0.99 \\
Batch size & 256 \\
\end{tabular}
\caption{Battle Experiment}
\end{table}

\section{Additional Evaluation}\label{appendix:additional-evaluation}
\begin{figure}[h!]
\captionsetup[subfigure]{skip=0pt, aboveskip=3pt}
    \begin{subfigure}[b]{0.33\linewidth}
        \centering
        \includegraphics[height=3.5cm]{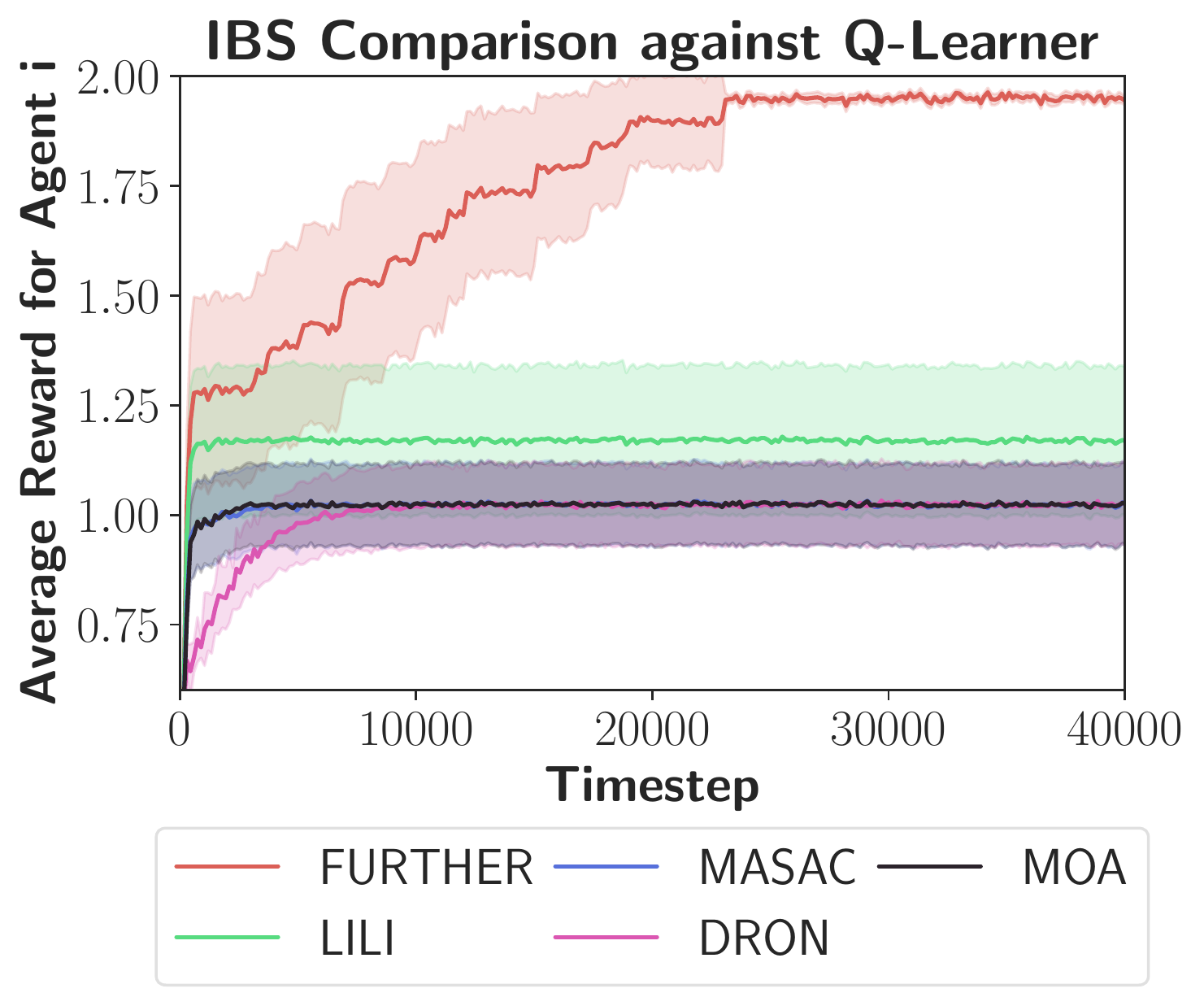}
        \caption{}
        \label{fig:ibs-result-appendix}
    \end{subfigure}
    \begin{subfigure}[b]{0.33\linewidth}
        \centering
        \includegraphics[height=3.5cm]{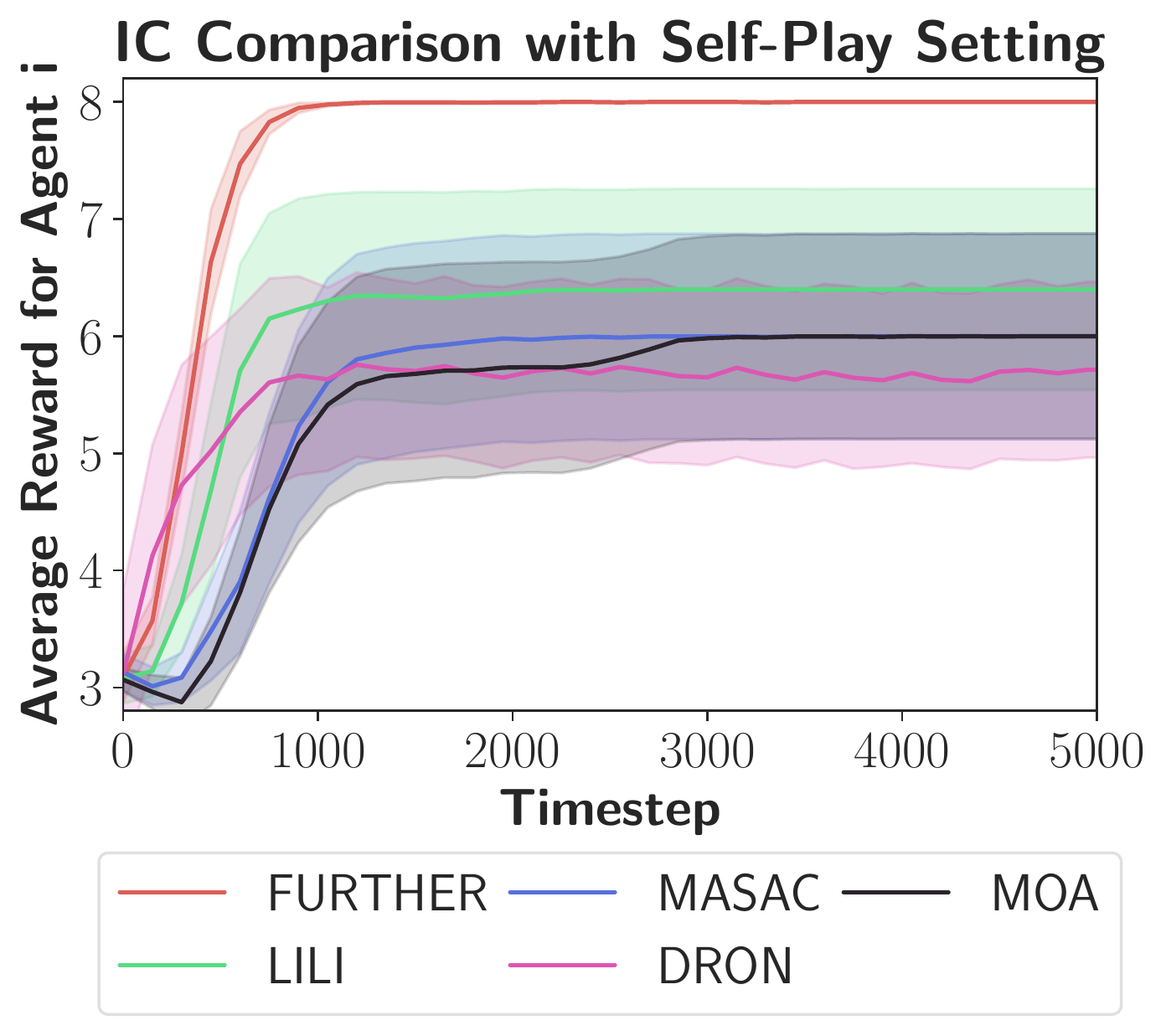}
        \caption{}
        \label{fig:ic-result-appendix}
    \end{subfigure}
    \begin{subfigure}[b]{0.33\linewidth}
        \centering
        \includegraphics[height=3.5cm]{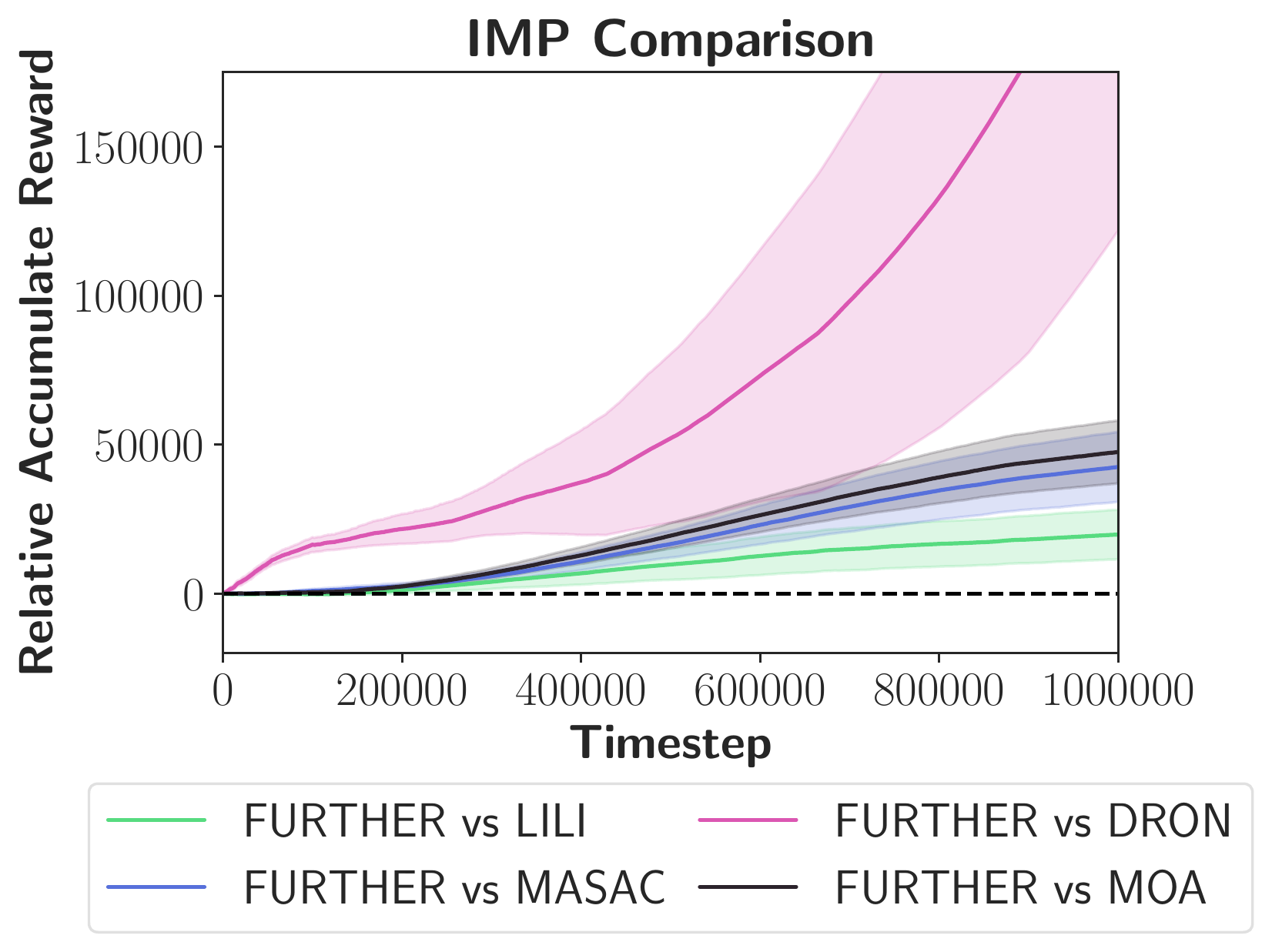}
        \caption{}
        \label{fig:imp-result-appendix}
    \end{subfigure}
    \vskip-0.09in
    \caption{\textbf{(a)} Convergence in IBS. The FURTHER agent achieves convergence to its optimal pure strategy Nash equilibrium. \textbf{(b)} Convergence in IC with self-play. The FURTHER team shows better converged performance than baselines. \textbf{(c)} A competitive play in IMP between FURTHER and baseline methods. FURTHER receives higher rewards than baselines over time.}
    \vskip-0.15in
\end{figure}
We show additional results about DRON and MOA in playing the iterated matrix games (see~\Cref{fig:ibs-result-appendix,fig:ic-result-appendix,fig:imp-result-appendix}). Because DRON and MOA also suffer from myopic evaluation, we generally observe the sub-optimal performance of these baselines in our evaluations. In particular, DRON does not consider the underlying learning of other agents, resulting in the FURTHER agent easily exploiting the DRON opponent in~\Cref{fig:imp-result-appendix}. We also observe that, while MOA's optimization of the influence reward can effectively learn coordination in sequential social dilemma domains \cite{leibo17socialdilemmas,jaques19social}, this influence reward optimization may not be useful in the competitive setting.

\section{Limitation and Societal Impact}\label{sec:limitation-and-impact}
FURTHER has a limitation that the framework does not consider an agent $i$'s own non-stationary policy. 
As discussed in~\Cref{sec:method}, it is ideal to maximize the average reward over the space of joint update functions, including $i$'s own update function. 
However, it is computationally intractable to solve long horizon meta-learning by considering $i$'s own policy dynamics, and this remains an active area of research~\citep{kim21metamapg,imaml,deleu2022continuous}.
Instead, we take a practical approach by assuming $i$'s fixed stationary policy. 
Taking an agent's own non-stationary policy into account is one of the future directions.
We also model the period as $k=1$ for simplicity in our experiments, and studying how varying $k$ has a potential effect on performance is another future direction.
Regarding the societal impact, while FURTHER can achieve a better social outcome in cooperative and self-play settings, a FURTHER agent aims to influence other agents to converge to desirable policies from its perspective. 
As such, there can be applications, where the framework may lead to negative societal impacts by taking advantage of other agents' defective decision-making.